\pdfoutput=1

\documentclass[10pt,twocolumn,letterpaper]{article}

\usepackage{cvpr}      

\usepackage{graphicx}
\usepackage{amsmath}
\usepackage{amssymb}
\usepackage{booktabs}

\usepackage{times}
\usepackage{epsfig}
\usepackage[noend]{algpseudocode}
\usepackage{algorithmicx,algorithm}
\usepackage{graphicx}
\usepackage{tabularray}
\usepackage{subfloat}
\usepackage{caption}
\usepackage{graphicx}
\usepackage{sidecap}
\usepackage{cuted}
\usepackage{multirow}
\usepackage{algorithm}
\usepackage{multicol}
\usepackage{cite}
\usepackage{bbm}
\usepackage{array}
\newcommand{\PreserveBackslash}[1]{\let\temp=\\#1\let\\=\temp}
\newcolumntype{C}[1]{>{\PreserveBackslash\centering}p{#1}}
\newcolumntype{R}[1]{>{\PreserveBackslash\raggedleft}p{#1}}
\newcolumntype{L}[1]{>{\PreserveBackslash\raggedright}p{#1}}

\usepackage[pagebackref,breaklinks,colorlinks]{hyperref}

\usepackage[capitalize]{cleveref}
\crefname{section}{Sec.}{Secs.}
\Crefname{section}{Section}{Sections}
\Crefname{table}{Table}{Tables}
\crefname{table}{Tab.}{Tabs.}

\def\cvprPaperID{224} 
\def\confName{CVPR}
\def\confYear{2022}

\begin{document}

\title{Rethinking Reconstruction Autoencoder-Based Out-of-Distribution Detection}

\author{Yibo Zhou\\
Beihang University\\
{\tt\small ybzhou@buaa.edu.cn}
\and
}
\maketitle

\begin{abstract}
In some scenarios, classifier requires detecting out-of-distribution samples far from its training data. With desirable characteristics, reconstruction autoencoder-based methods deal with this problem by using input reconstruction error as a metric of novelty vs. normality. We formulate the essence of such approach as a quadruplet domain translation with an intrinsic bias to only query for a proxy of conditional data uncertainty. Accordingly, an improvement direction is formalized as maximumly compressing the autoencoder's latent space while ensuring its reconstructive power for acting as a described domain translator. From it, strategies are introduced including semantic reconstruction, data certainty decomposition and normalized L2 distance to substantially improve original methods, which together establish state-of-the-art performance on various benchmarks, \eg, the FPR@95\%TPR of CIFAR-100 vs. TinyImagenet-crop on Wide-ResNet is 0.2\%. Importantly, our method works without any additional data, hard-to-implement structure, time-consuming pipeline, and even harming the classification accuracy of known classes. Code has been released on github.

\end{abstract}
\section{Introduction}
Supervised discriminative deep classifiers are de facto designed under a static closed-world assumption where the data, which model face in the deployment environment, is supposed to be sampled from the same distribution as training set \cite{hsu2020generalized}. However, for applications in the wild, such as the safety-critical autonomous vehicles, test data is often hard to be known a priori. Worse, given that neural networks especially those built on relu or softmax can easily produce not only erroneous, but arbitrarily confident so, class predictions even for totally unrecognizable or irrelevant samples \cite{hein2019relu,hendrycks2016baseline,nguyen2015deep,goodfellow2014explaining}, it is consequential to empower system to flag or abstain predictions from unkowns.

Out-of-distribution (OoD) detection is a binary classification of detecting inputs sampled from distribution different from training data\cite{hendrycks2016baseline}. Many existing methods rely on training or tuning with data labelled as OoD from other categories\cite{Thulasidasan2021AnEB,yu2019unsupervised}, adversaries\cite{malinin2019reverse,lee2018simple} or the leave-out subset of training samples\cite{vyas2018out}. However, it is intractable to cover the full space of OoD particularly for data with large dimensions (\eg, an image)\cite{1969Some}, resulting in that a supervised method capturing limited facets of OoD distribution hardly generalizes without a data selection bias\cite{2018A}. Meanwhile, the classifier's accuracy on in-distribution (ID) images would be fluctuated by introducing OoD data with additional training objectives \cite{vyas2018out, Thulasidasan2021AnEB}. Such factors expose the unsupervised nature of OoD detection.

In a class of unsupervised approaches, input reconstruction residual is considered a novelty metric to avoid the above problems\cite{pimentel2014review}. The essential assumption is that an autoencoder learned to rebuild ID samples could not reconstruct OoD comparatively during testing\cite{2019Deep}. However, it has been extensively reported that autoencoders can effectively rebuild kinds of OoD samples even better \cite{denouden2018improving,pimentel2014review}, causing an inferior performance of such methods when applied to challenging multi-class OoD detection tasks. In this paper, we investigate in a precise manner the problematic reconstruction of outliers from the perspective of quadruplet domain translation by formalizing two concrete preconditions under which the reconstruction error of an input is a valid data uncertainty measure. First, its latent feature lies within the domain of encoded ID samples. Second, the decoder has adequate reconstructive power to bridge the domains of ID images and their latent representations. 

Precondition 1 requires system to capture the outliers of latent representations. Without relying on kernel density estimators\cite{1956Remarks,denouden2018improving,2017Joint} performing weakly in high dimensions\cite{2020Roundtrip}, similar to the scheme considering for compact latent representations \cite{J2016End,2018Latent}, we minimize a regularization loss to restrict ID latent features distributed within a certain space. In conjunction with the training of input reconstruction, it is explicated that, when this space is compressed sufficiently, any latent feature outlier will lie outside it. 

There is a coupled problem that an over-restricted latent space might not provide sufficient reconstructive power for large-scale ID images, which violates precondition 2. To mitigate it, we change the reconstruction target from the image to its extracted activation vector (AV) feature to reduce the unnecessary requirement of the expressiveness of latent space. As a further step, from the perspective of domain translation, we deduce a base equation to model the inherent connection between input reconstruction error and data certainty. By the probability chain rule, we further factorize it to express data certainty as a product of conditional densities defined by the layer-by-layer encoder feature reconstruction error. It is proved that, albeit the considerable information loss accumulated from the whole encoding process making it hard to recover input directly from a compact latent space to satisfy precondition 2, there exists an equivalent precondition feasibly to be met as it only requires to recover the information lost after each irreversible encoding layer respectively. Consequently, inspired by above concepts, we present a theoretically well-defined framework for OoD detection, namely layerwise semantic reconstruction.

In this framework, we employ only one fully connected (FC) layer and softmax function as the encoder architecture and leverage simple cross entropy loss to restrict the latent space. We provide both experimental evidences and mathematical insights, indicating that under such a setting the maximum value of a latent feature can be utilized as a domain affinity for filtering out OoD data potentially to be reconstructed. Since classifier is inclined to produce smaller neural activations on OoD data\cite{shalev2018out,taigman2015web}, to render our method robust against it, reconstruction accuracy is evaluated with the proposed normalized L2 distance.

Our contribution is three-fold: First, we establish a novel perspective for understanding autoencoder-based OoD detections, figuring out one direction to improve them by maximumly condensing autoencoder’s latent space while reserving sufficient reconstructive power on ID data. Rooting from it, second, a framework of layerwise semantic reconstruction is developed. Third, along with ablation and robustness studies, we provide comprehensive analysis of our proposal using various benchmark datasets to demonstrate its efficacy, indicating that the potentiality of autoencoder-based methods is not as bleak as previously displayed.

\begin{itemize}
\item Our method delivers comparable performance to SOTA methods on various challenging benchmarks.
\item As an auxiliary module, our OoD detector is trained in a way orthogonal to the classifier.
\item Operated in an unsupervised mode of efficiency and applicability, our method requires no additional data.
\end{itemize}
\section{Background and Related Work} \label{backrelat}

An autoencoder consists of an encoder to project an input into a latent space with fewer dimensions (also known as bottleneck feature) and a decoder to recover the input from its latent representation. During training, input is processed sequentially through the encoder and decoder to minimize the reconstruction error which is a discrepancy like L2 distance between the input and the reconstruction output from the decoder. The basic assumption in the reconstruction autoencoder-based OoD detector is that an autoencoder trained exclusively to recover ID samples could not succeed in reconstructing OoD samples. Hence, reconstruction error becomes a potentially effective decision function of OoD detection in testing. However, contrary to this expectation, autoencoders are reported to accurately reconstruct different types of OoD samples \cite{2016Outlier, 2020Fixing}.

It was pointed out in \cite{2017Robust} that the gross noise within training samples makes it hard to learn a robust latent representation of the majority of ID samples. Hence, a neural network extension of robust principal component analysis (RPCA) \cite{2010Robust} was proposed. In \cite{denouden2018improving} the OoD detection methods of Mahalanobis distance and autoencoder were merged into a unified framework, supposing that the latter could be thus enhanced. Recent work on latent space autoregression \cite{2018Latent} proposed to constrain the autoencoder from an identity function by applying an autoregressive density estimator to minimize the differential entropy of the distribution of ID latent features. Our proposal also takes advantage of restricting the distribution of ID latent features. Differently, we seek a maximumly compressed latent space that tightly covers the the domain of ID latent codes to get an approximation of this domain without parametric density estimator\cite{2018Deep} . The comparison of both methods is detailed in the experiment section, showing that our method outperforms the existing method by considerable margins.

\section{Hypothesis} \label{main_concept}
In this section, we introduce the concepts of our framework. Formally, it is supposed that there is a domain $\mathcal X \subset \mathbb{R}^{N}$ characterized by all of the ID images. We consider a sample $\boldsymbol z$ as an OoD sample when it is not in $\mathcal X$. In essence, any $\boldsymbol x \in \mathcal X$ is an ID sample and any $\boldsymbol z \in \mathcal Z = \complement_{\mathbb{R}^{N}}\mathcal X$ is an OoD. For a trained autoencoder consisting of an encoder $E(\cdot)$ and a decoder $D(\cdot)$ on a dataset of ID samples, let the domain spanned by $E(\boldsymbol x),\forall \boldsymbol x \in \mathcal X$ be $\mathcal S_{ID} \subset \mathbb{R}^{M}$. Under the hypothesis that \emph{this decoder has sufficient reconstructive power to act as a domain translator of $\mathcal S_{ID} \mapsto \mathcal X$}, $E(\cdot)$ and $D(\cdot)$ are functions learning the forward and backward mapping relationships between $\mathcal X$ and $\mathcal S_{ID}$:
\begin{equation}
E(\boldsymbol x) \in \mathcal S_{ID}, \forall \boldsymbol x \in \mathcal X \label{con:encoder},
\end{equation} \vspace{-4.0ex} 
\begin{equation}
D(\boldsymbol f) \in \mathcal X, \forall \boldsymbol f \in \mathcal S_{ID}\label{decoder}.
\end{equation}

Since $\mathcal X$ and $\mathcal Z$ might overlap when encoded, if we represent the domain of encoded OoD samples as $\mathcal S_{OoD}$, it could be decomposed as $\mathcal S_{OoD}=\mathcal S_{OoD} \cap \mathcal S_{ID} + \mathcal S_{OoD} \cap \complement_{\mathbb{R}^{M}}\mathcal S_{ID}$. Using Eq.\ref{decoder}, any OoD sample encoded into $\mathcal S_{ID} \cap \mathcal S_{OoD}$ can be decoded back to $\mathcal X$. For this part of OoD samples, we expect a larger reconstruction error since they are decoded into a different domain $\mathcal X$. Also, for these OoD samples, the further they lie from $\mathcal X$ in $\mathbb{R}^{N}$, the larger the reconstruction error might be. This is consistent with the characteristic of a measure of data certainty. 

\begin{figure}[t]
\centering
\includegraphics[height=4.5cm,width=6.55cm]{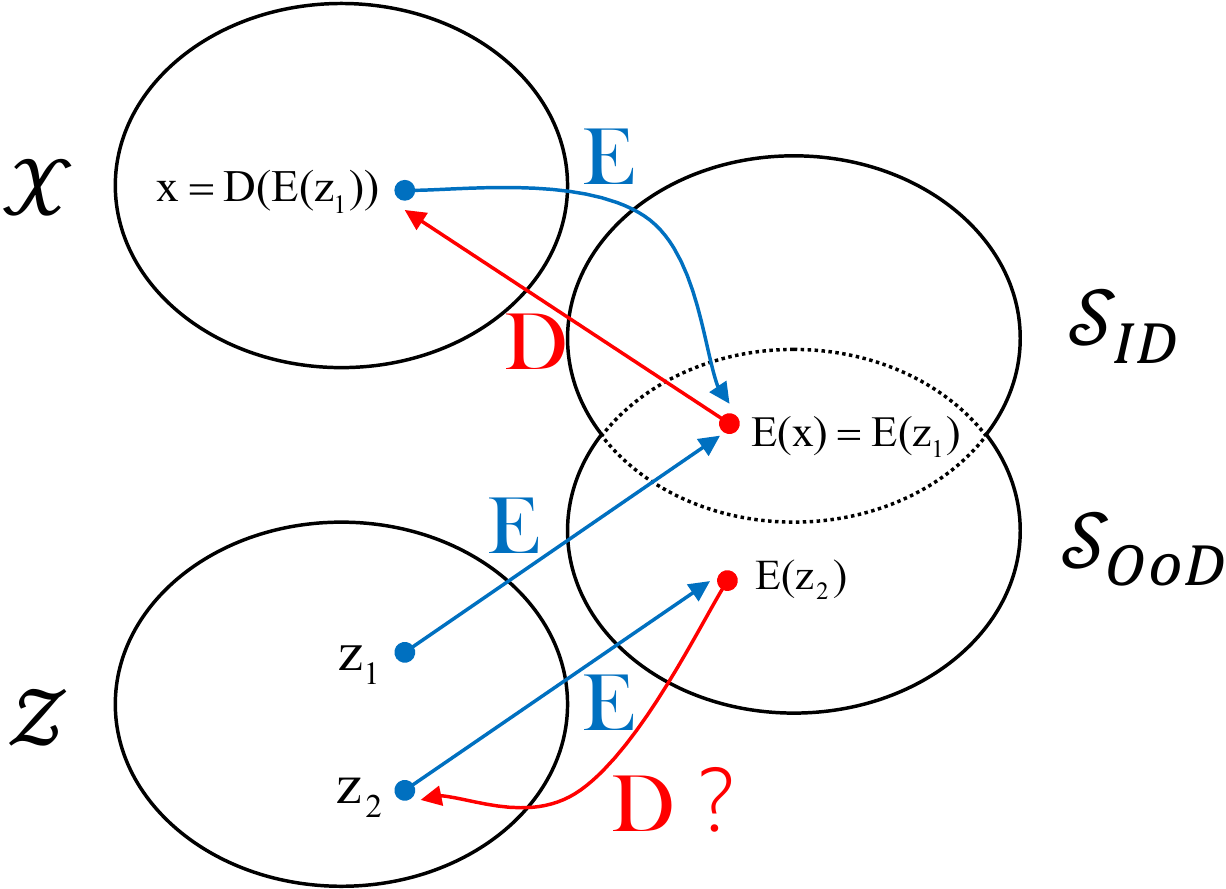}
\caption{Illustration of the described quadruplet domain translation. For an OoD sample $z_1$ encoded into $\mathcal S_{ID} \cap \mathcal S_{OoD}$, its latent representation $E(z_1)$ is equal potentially to that of an ID sample $x$. Therefore, $E(z_1)$ can be decoded to a different sample $x$ within $\mathcal X$, resulting in a large reconstruction error. However, for an OoD sample $z_2$ with latent representation $E(z_2)$ lying outside $\mathcal S_{ID}$, it offers no guarantee that it could not be reconstructed well.}
\label{bconcept}
\end{figure}

However, for an OoD sample $\boldsymbol z$ with latent feature $E(\boldsymbol z)\\ \in \mathcal S_{OoD} \cap \complement_{\mathbb{R}^{M}}\mathcal S_{ID}$, it is not guaranteed to prevent the reconstruction of it. Although $E(\boldsymbol z)$ can be far from the encoded ID data, $\boldsymbol z$ might lie on a manifold determined by the parameters and architecture of the autoencoder, on which data could be precisely reconstructed. An example is an OoD image of all zeros. After being processed by an autoencoder without shift operations, its reconstruction would be naturally all zeros, resulting in a perfect reconstruction of OoD. Naturally encapsulated with concept graphed in Fig.\ref{bconcept}, \emph{for an agnostic input $\boldsymbol r$, given $E(\boldsymbol r) \in \mathcal S_{ID}$, its reconstruction error is a valid measurement of data uncertainty $P (\boldsymbol r \notin \mathcal X)$}. This concept is modeled as 

\begin{equation}
P(\boldsymbol r \in \mathcal X | E(\boldsymbol r) \in \mathcal S_{ID}) = F(\emph{Dist}(\boldsymbol r, D(E(\boldsymbol r)))), \label{essence}
\end{equation} 

\noindent for a decoder $D(\cdot)$ satisfying Eq.\ref{decoder}, an appropriate distance metric \emph{Dist($\cdot$,$\cdot$)} applied to compute reconstruction error and a certain monotonically decreasing function $F(\cdot):\mathbb{R} \mapsto [0,1]$ to map reconstruction error into probability. Noticing $P(\boldsymbol r \in \mathcal X) = P (\boldsymbol r \in \mathcal X, E(r) \in \mathcal S_{ID}) + P(\boldsymbol r \in \mathcal X, E(\boldsymbol r) \notin \mathcal S_{ID})$, by the definition of $\mathcal S_{ID}$ we have $P(\boldsymbol r \in \mathcal X, E(\boldsymbol r) \notin \mathcal S_{ID}) = 0$, and therefore 

 \begin{small}
 \begin{equation}
 \begin{aligned}
 \begin{split}
 P(\boldsymbol r \in \mathcal X)& = P (\boldsymbol r \in \mathcal X, E(\boldsymbol r) \in \mathcal S_{ID}) + 0 \\ 
 & = P(\boldsymbol r \in \mathcal X | E(\boldsymbol r) \in \mathcal S_{ID}) \cdot P(E(\boldsymbol r) \in \mathcal S_{ID}) \\
 & = F(\emph{Dist}(\boldsymbol r,D(E(\boldsymbol r)))) \cdot P(E(\boldsymbol r) \in \mathcal S_{ID}). \label{func} 
  \end{split}
  \end{aligned}
  \end{equation}
  \end{small}
 
\begin{figure}[t]
\centering
\includegraphics[height=3.8cm,width=8.8cm]{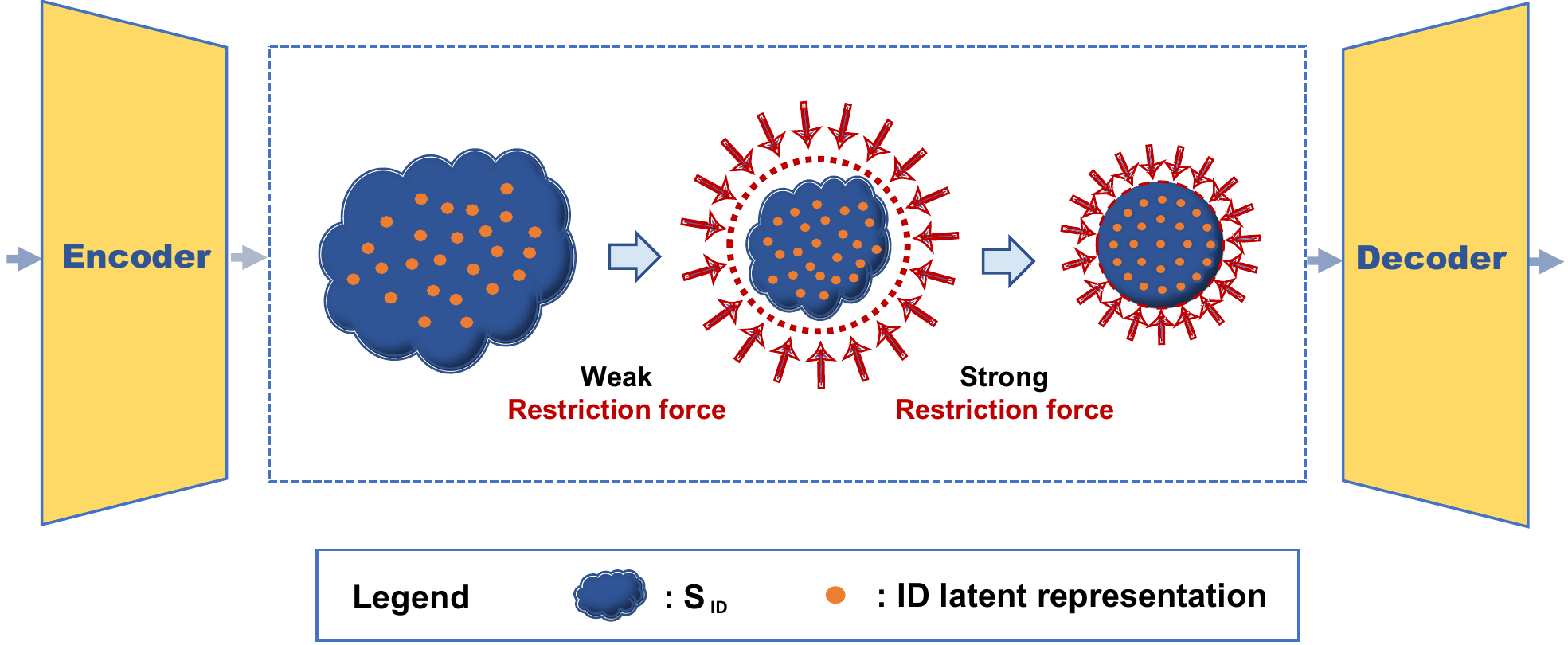}
\caption{Illustration of the transition of $\mathcal S_{ID}$ when the restrictive power imposed over latent codes increases. During training, any deviation of the latent codes from the latent space (red-dot circle) would be penalized greatly. By tightening this space sufficiently, in principle it would be mostly utilized to satisfy the jointly learned reconstruction task. Thus, detecting the outlier of $\mathcal S_{ID}$ is approximately equal to identifying the feature outside this latent space.}
\label{transition}
\end{figure}

To estimate $P(E(\boldsymbol r) \in \mathcal S_{ID})$, instead of fitting $E(\boldsymbol x) \in \mathcal S_{ID}$ extracted in a trained autoencoder with specific family of distributions like Gaussian to get a biased density estimator under restrictive assumption, we could apply an additional regularization loss on ID latent features in the training stage to force them to reside within a known compact space for a better approximation of $\mathcal S_{ID}$. This idea is represented in Fig.\ref{transition}. Since $\mathcal S_{ID}$ is covered by this restricted latent space, with the increasing restrictive power imposed over ID latent features, to retain the power of the jointly learned reconstruction task, the model implicitly seek for a larger use of this latent space. It means that a restricted latent space could be more filled by $\mathcal S_{ID}$. Therefore, estimating $P(E(\boldsymbol r) \in \mathcal S_{ID})$ is much equal to estimating the probability that $E(\boldsymbol r)$ is in this restricted latent space.

However, this regularizer in the OoD detection setting is coupled simultaneously with the problem that an over-limited latent space can result in reconstruction difficulty in ID images. It potentially violates the fundamental assumption that $D(\cdot)$ acts as a domain translator of $\mathcal S_{ID} \mapsto \mathcal X$, pushing $F(\emph{Dist}(\boldsymbol r,D(E(\boldsymbol r))))$ away from an accurate estimate of $P(\boldsymbol r \in \mathcal X | E(\boldsymbol r) \in \mathcal S_{ID})$. Thus, the main interest here is to maximumly condense the autoencoder's latent space whilst retaining its power to reconstruct the ID data. 

Equipped with this perspective, images especially those with a large spatial resolution are not good selections for reconstruction target. Satisfying Eq.\ref{decoder} for image requires precisely recovering its myriad details irrelevant to OoD detection. Therefore, the unnecessary requirement rises for data collection, computation, and even worse, the size of latent space which is not desirable to estimate $P(E(\boldsymbol r) \in \mathcal S_{ID})$. 

Since the higher-level feature extracted in a classifier can be considered as a learned representation acquiring almost all the information about the original image except the less class-representative details \cite{2015Inverting} and the existence of a trained classifier is often implied in this topic, to avoid the complexity of reconstruction on pixel space, we turn to reconstruct AV feature which is the output at penultimate layer of a classifier as the low dimensional semantic representation of image to quantify data normality. As argued in \cite{yoshihashi2019classification}, image-based detector is sensitive to image scale. In contrast, AVs can be set as fixed size and only keep the most task-relevant information, relieving the susceptibility of original methods to the scale of image. Also, an advantage is that it does not require the input images and access to the trained model, having implications for privacy. In the following, we will denote the AV feature of an agnostic image as $\boldsymbol v$ and the domain of AV features extracted from ID images as $\mathcal V \subset \mathbb{R}^{H}$, where $H$ represents the dimensionality of $\boldsymbol v$. 

As the latent space is compressed constantly, the part of input information lost in it increases. Noticing that this unrecoverable part of information is accumulated from every single irreversible layer of the encoder, we factorize $P(\boldsymbol v \in \mathcal V) $ as follows, to sidestep the difficulty of recovering input directly from a latent space lacking of expressiveness: for an encoder $E(\cdot)$ with $n$ layers, we represent $f_{i}(\boldsymbol v),i = 1,…,n$ as the output feature at $i^{th}$ layer given an input AV $\boldsymbol v$. If we denote the domain of $f_{i}(\boldsymbol v),\boldsymbol  v \in \mathcal V$ as $\mathcal F_{i}$, naturally $P(f_1(\boldsymbol v) \in \mathcal F_1,f_{2}(\boldsymbol v) \in \mathcal F_{2},...,f_{n}(\boldsymbol v) \in \mathcal F_n |\boldsymbol  v \in \mathcal V) = 1$. Therefore $P(\boldsymbol v \in \mathcal V) = P(\boldsymbol v \in \mathcal V) \cdot 1= P(\boldsymbol v \in \mathcal V) \cdot P(f_1(\boldsymbol v) \in \mathcal F_1,f_{2}(\boldsymbol v) \in \mathcal F_{2},...,f_{n}(\boldsymbol v) \in \mathcal F_n |\boldsymbol  v \in \mathcal V) = P(\boldsymbol v \in \mathcal V, f_1(\boldsymbol v) \in \mathcal F_1, f_2(\boldsymbol v) \in \mathcal F_2,...,f_n(\boldsymbol v) \in \mathcal F_n)$. Assuming that the forward propagation of this encoder forms a Markov process of $\boldsymbol v \stackrel{layer1}{\longrightarrow} f_1(\boldsymbol v) \stackrel{layer2}{\longrightarrow} f_2(\boldsymbol v) \stackrel{layer3}{\longrightarrow} ... \stackrel{layern}{\longrightarrow} f_n(\boldsymbol v)$, any feature can be determined by any of its preceding feature. Hence, $P(f_i(\boldsymbol v) \in \mathcal F_i,f_{i+1} (\boldsymbol v) \in \mathcal F_{i+1},...,f_n(\boldsymbol v) \in\mathcal F_n) = P(f_i(\boldsymbol v) \in \mathcal F_i)$, then we have

 \begin{small}
 \begin{equation}
 \begin{aligned}
P(\boldsymbol v \in &\mathcal V) = P(\boldsymbol v \in \mathcal V, f_1(\boldsymbol v) \in \mathcal F_1, f_2(\boldsymbol v) \in \mathcal F_2,...,f_n(\boldsymbol v) \in \mathcal F_n) \\ 
& = P(f_n(\boldsymbol v) \in \mathcal F_n) \cdot P (\boldsymbol v \in \mathcal V | f_1(\boldsymbol v) \in \mathcal F_1,...,f_n(\boldsymbol v) \in \mathcal F_n) \\
& \cdot \prod_{i=1}^{n-1}P(f_i(\boldsymbol v) \in \mathcal F_i | f_{i+1} (\boldsymbol v) \in \mathcal F_{i+1},...,f_n(\boldsymbol v) \in \mathcal F_n) \\ 
& = P(f_n(\boldsymbol v) \in \mathcal F_n) \cdot P(\boldsymbol v \in \mathcal V | f_1(\boldsymbol v) \in \mathcal F_1)\\
& \cdot \prod_{i=1}^{n-1}P(f_i(\boldsymbol v) \in \mathcal F_i | f_{i+1} (\boldsymbol v) \in \mathcal F_{i+1}). \\  \nonumber
 \end{aligned}
 \end{equation}
 \end{small} \vspace{-2.5ex} 

Extended from Eq.\ref{essence}, $F(\emph{Dist}(f_i(\boldsymbol v), D_{i+1}(f_{i+1}(\boldsymbol v)))$ can be utilized as an estimation of $P(f_i(\boldsymbol v) \in \mathcal F_i | f_{i+1}(\boldsymbol v) \in \mathcal F_{i+1})$ for $i = 1,...,n-1$, where $D_{i+1}$ is a decoder capable of recovering $f_i(\boldsymbol v)$ from $f_{i+1}(\boldsymbol v)$, $\forall \boldsymbol v \in \mathcal V$.  After substitution, we get formally

 \begin{small}
 \begin{equation}
 \begin{aligned}
P(\boldsymbol v \in \mathcal V)& = P(f_n(\boldsymbol v) \in \mathcal F_n) \cdot F(\emph{Dist}(\boldsymbol v,D_1(f_1(\boldsymbol v)))) \\
&\cdot \prod_{i=1}^{n-1}F(\emph{Dist}(f_i(\boldsymbol v),D_{i+1}(f_{i+1}(\boldsymbol v)))). \label{input}
 \end{aligned}
 \end{equation}
 \end{small} 
 
In contrast to Eq.\ref{func}, this reformulation in Eq.\ref{input} allows us to approximate $P(\boldsymbol v \in \mathcal V)$ without direct input reconstruction from a restricted latent space, which requires recovering the considerable accumulated information loss at one round. Instead, it could be accomplished through a series of decoders $D_i$ focusing on recovering the information lost after each single encoding layer, which is only a smaller divided part of the total information loss and therefore relatively easy to be recovered. Since the reconstructive ability of a decoder is closely related to its performance as an estimator of the defined conditional probability, Eq.\ref{input} allows a better estimation of input certainty in practice, particularly when the latent space is over restricted.

\section{Practical Approach}

After establishing the framework, we must specify the form of $\emph{Dist}(\cdot,\cdot)$, $F(\cdot)$ and the regularizer. Here we give our simple but effective designs in order.

By minimizing a reconstruction discrepancy like $L2$ distance, we can train the decoders in Eq.\ref{input} (assuming that the reconstruction error follows a Gaussian distribution). However, such a norm-correlated distance metric is not appropriate for use as $\emph{Dist}(\cdot,\cdot)$ in test time. Obviously, feature with smaller L2-norm tends to have smaller $L2$ reconstruction loss. In order to set the theoretical ground for supporting this intuition, we discuss and prove the fact that an upper bound of input's L2 reconstruction error is definite and is approximately proportional to its L2-norm. 

Since the input of our autoencoder is one-dimensional AV feature, it is natural to employ FC network as the architecture of the encoder and decoder. Noticeably, the convolutional layer, shortcut connection and average pooling are all in essence linear mappings, and therefore the additional use of them will not concern the following discussion. For a FC autoencoder with $L$ layers, we denote the weight matrix of $l^{th}$ FC layer as $W^{l} \in  \mathbb{R}^{n^l \times n^{l-1}}$, the offset bias as $\boldsymbol b^{l} \in \mathbb{R}^{n^l}$, the activation function as $\sigma(\cdot)$ and the input as $\boldsymbol x \in \mathbb{R}^{H}$. Then the pre-activation output of $l^{th}$ layer could be recursively written as

 \begin{equation}
 \begin{aligned}
 f^{l}(\boldsymbol x) = W^{l}\sigma(f^{l-1}(\boldsymbol x))+\boldsymbol b^{l}. \nonumber
 \end{aligned}
 \end{equation} 

Although activation functions in the context of deep learning are always nonlinear in the full space, they could be approximately considered as linear in a certain polytope. For instances, given an input in $\mathbb{R}^{1}$, relu is linear if only the region of $[0,+\infty)$ or $(-\infty, 0)$ is considered. Sigmoid could be approximated as linear exclusively in the region of $(-\infty, -\alpha)$,  $[-\alpha, \alpha)$ or $[\alpha, +\infty)$, for some $\alpha > 0$. For simplicity, in the following discussion we consider relu as the applied activation function $\sigma(\cdot)$. Similar to \cite{hein2019relu}, our FC autoencoder can be expressed as a piecewise affine function

 \begin{equation}
 \begin{aligned}
 f^{L}(\boldsymbol x) = & W^{L}\sigma(W^{L-1}\sigma(\\
 &...\sigma(W^{1}\boldsymbol x+\boldsymbol b^{1})...)+\boldsymbol b^{L-1})+\boldsymbol b^{L}\\
 = &W^{L}\Lambda^{L-1}(\boldsymbol x)(W^{L-1}\Lambda^{L-2}(\boldsymbol x)(\\
 &...\Lambda^{1}(\boldsymbol x)(W^{1} \boldsymbol x+\boldsymbol b^{1})....)+\boldsymbol b^{L-1})+\boldsymbol b^{L} \\
 = &\Gamma \boldsymbol{x} + B, \nonumber
 \label{decom}
 \end{aligned}
 \end{equation} 

\noindent where $\Lambda^{l}(\boldsymbol x) \in \mathbb{R}^{n^l \times n^{l}}$, for $l = 1,...,L-1$ are diagonal matrices defined as  

\begin{equation}
\Lambda^{l}(\boldsymbol x) = 
\begin{bmatrix}
\mathbbm{1}(f^{l}_1(\boldsymbol x) > 0) &   &  \\
& ... &\\
&  &\mathbbm{1}(f^{l}_{n^{l}}(\boldsymbol x) > 0) \nonumber
\end{bmatrix},
\end{equation}

\noindent and $\Gamma \in \mathbb{R}^{H \times H}$ and $B \in \mathbb{R}^{H}$ are matrices defined as
 
\begin{equation}
\begin{aligned}
&\Gamma = W^{L}(\prod_{i=1}^{L-1}\Lambda^{L-i}(\boldsymbol x)W^{L-i}), \\
&B = \sum_{i=1}^{L-1}(\prod_{k=1}^{L-i}W^{L+1-k}\Lambda^{L-k}(\boldsymbol x))\boldsymbol b^{i} + \boldsymbol b^{L}. \nonumber
\end{aligned}
\end{equation}

\noindent We can further have

\begin{equation}
\begin{aligned}
\Vert \boldsymbol x -  f^{L}(\boldsymbol x)\Vert& = \Vert \boldsymbol x -  \Gamma \boldsymbol x - B \Vert \\
&\le \Vert \boldsymbol x -  \Gamma \boldsymbol x \Vert + \Vert B \Vert \\
&\le \Vert I - \Gamma \Vert \Vert \boldsymbol x \Vert + \Vert B \Vert. \nonumber
\end{aligned}
\end{equation}

The number of total possible variants of $\Gamma$ and $B$ is $2^{\sum_{i=1}^{L-1}n^{i}}$, and the specific forms of $\Gamma$ and $B$ are determined by which polytope defined as intersection of $\sum_{i=1}^{L-1}n^{i}$ half spaces in $\mathbb{R}^{H}$ the input $\boldsymbol x$ is in. Therefore, given an input, an upper bound of its L2 reconstruction error is definite, and is approximately proportional to its norm. This result clearly supports our claim that \emph{feature with smaller norm tends to have smaller $L1$ or $L2$ reconstruction loss}. For instances, given a feature with norm close to zero, its reconstruction error is approximately $\Vert B \Vert$. While for a feature with norm $\rightarrow \infty$, its reconstruction error tends to be arbitrarily larger than $\Vert B \Vert$. 

Unfortunately, the neural network is inclined to produce smaller neural activations on inputs that model is not familiar with \cite{shalev2018out,taigman2015web}, \ie, OoD. As a result, the normality measure in Eq.\ref{input} with $L1$ or $L2$ distance applied to evaluate reconstruction accuracy can produce a relatively larger value for OoD (see sec.\ref{threshold}). This is opposite to what we expected. Thus, we propose the normalized L2 distance (NL2) as the specific form of $\emph{Dist}(\cdot,\cdot)$ to compute residuals 

 \begin{equation}
 \begin{aligned}
 \emph{Dist}(\boldsymbol f,\tilde{\boldsymbol f}) = \emph{NL2}(\boldsymbol f,\tilde{\boldsymbol f}) = \Vert \frac{\boldsymbol f}{\Vert \boldsymbol f \Vert} - \frac{\tilde{\boldsymbol f}}{\Vert \boldsymbol f \Vert} \Vert, \label{normalized} 
 \end{aligned}
 \end{equation}

\noindent where $\tilde{\boldsymbol f}$ represents the reconstruction of $\boldsymbol f$. It should be noted that $\tilde{\boldsymbol f}$ is normalized with $\Vert \boldsymbol f \Vert$ instead of the norm of itself. With NL2 distance, the reconstruction is assessed between the projection of $\boldsymbol f$ on the surface of a unit hypersphere and its equally scaled reconstruction. Therefore, the negative influence of the feature norm is eliminated.

$F(\cdot)$ translates the reconstruction error into a conditional probability that can be measured as the inlier reconstruction error belongingness from previous discussion. Here, we interpret this belongingness as, for a given feature, the probability that its reconstruction error is smaller than that of an ID image‘s feature, which is naturally the complementary cumulative density function (CCDF) of the reconstruction error of ID features. Assuming that the reconstruction error of ID data distributes as Gaussian, we have \vspace{-2.0ex} 

 \begin{equation}
 \begin{aligned}
&F(\emph{NL2}(\boldsymbol f,\tilde{\boldsymbol f})) = \Psi(\emph{NL2}(\boldsymbol f,\tilde{\boldsymbol f}) | \mu,\sigma + \epsilon), \label{GCCDF}
 \end{aligned}
 \end{equation} 

\noindent in which $\Psi(\cdot )$ is the CCDF of a Gaussian and $\mu$ and $\sigma$ are the parameters of Gaussian derived from validation data. Since a minor decrease leads to a substantial change of a value's numerical magnitude when it is close to 0, and our normality measure in Eq.\ref{input} is a product of multiple terms of Eq.\ref{GCCDF} within the range of $0$ to $1$, with increasing the number of encoder layers, the final estimation can collapse to zero rapidly even for training data. To avoid this situation, we introduce additive term $\epsilon$ in Eq.\ref{GCCDF} to prevent it from producing values smaller than 0.1 on features of ID validation data.

  \begin{algorithm}[t]
        \setlength{\textfloatsep}{0pt}
	\caption{Training pipeline}  
	\begin{algorithmic}[1] 
	         \Require ID training set: \{($\boldsymbol x_i,y_i)\}_{i=1}^{k}$, and ID validation set: \{($\boldsymbol x_i,y_i)\}_{i=k+1}^{n}$
		\Require Network $M(\cdot)$ fully trained on ID training set for classification of ID classes
		\State Freeze all the parameters of network $M(\cdot)$, and jointly train $W \in \mathbb{R}^{H \times C}$($C$ is the number of ID classes) and two decoders $D_1 \& D_2$ to minimize the loss $\mathcal{L}$ \label{training_p} \vspace{-1ex} 
		\Statex   \begin{footnotesize} \begin{equation}  \begin{aligned} \mathcal{L} = \mathcal{L}_1 + \mathcal{L}_2 + \lambda \cdot \mathcal{L}_{regularizer} \nonumber \end{aligned} \end{equation} \end{footnotesize} \vspace{-6ex} 
		\Statex   \begin{footnotesize} \begin{equation}  \begin{aligned} \mathcal{L}_1 = \sum_{i=1}^k \Vert \boldsymbol v_i - D_1(W \boldsymbol v_i) \Vert , \mathcal{L}_2 = \sum_{i=1}^k \Vert \frac{W \boldsymbol v_i}{T} - D_2(\emph{S}(\frac{W \boldsymbol v_i}{T})) \Vert \nonumber \end{aligned} \end{equation} \end{footnotesize} \vspace{-4.5ex} 
		\Statex   \begin{footnotesize} \begin{equation}  \begin{aligned} \mathcal{L}_{regularizer} = - \sum_{i=1}^k \sum_{j=1}^C \mathbbm{1}(j = y_i) \log \emph{S}(W \boldsymbol v_i)_j, \nonumber \end{aligned} \end{equation} \end{footnotesize}
		\Statex where $\boldsymbol v_i$ is $\boldsymbol x_i$'s AV feature extracted in $M(\cdot)$ and $\lambda$ is the weight of regularization loss
		\State After training, compute: \vspace{-2ex} 
		\Statex \begin{footnotesize} \begin{equation}  \begin{aligned} (\mu_0,\sigma_0) = norm.fit(\{\emph{S}(\frac{W \boldsymbol v_{i}}{T})_{\overline y_{i}}\}_{i=k+1}^{n})\nonumber \end{aligned}\end{equation} \end{footnotesize} \vspace{-3.5ex} 
		\Statex \begin{footnotesize} \begin{equation}  \begin{aligned} (\mu_1,\sigma_1) = norm.fit(\{\Vert \frac{\boldsymbol v_i}{\Vert \boldsymbol v_i \Vert} - \frac{D_1(W \boldsymbol v_i)}{\Vert \boldsymbol v_i \Vert} \Vert\}_{i=k+1}^{n})\nonumber \end{aligned}\end{equation} \end{footnotesize} \vspace{-4.2ex} 
		\Statex \begin{footnotesize} \begin{equation}  \begin{aligned} (\mu_2,\sigma_2) = norm.fit(\{\Vert \frac{W \boldsymbol v_i}{\Vert W \boldsymbol v_i \Vert} - \frac{D_2(\emph{S}(\frac{W \boldsymbol v_i}{T}))}{\Vert \frac{W \boldsymbol v_i}{T} \Vert} \Vert\}_{i=k+1}^{n})\nonumber \end{aligned}\end{equation} \end{footnotesize} \vspace{-2ex} 
		\State \Return $D_1,D_2,W,(\mu_0,\sigma_0),(\mu_1,\sigma_1)\;and\;(\mu_2,\sigma_2)$
	\end{algorithmic}  
\end{algorithm} 

For an accurate estimate of  $P(E(\boldsymbol v) \in \mathcal S_{ID})$, \ie, $P(f_n\\(\boldsymbol v) \in \mathcal F_n)$ in Eq.\ref{input}, it is essential to encode ID AVs into a depictable compact space closely covering $\mathcal S_{ID}$. Since a classifier $M(\cdot)$ was optimized to translate ID images into AV features capable of being converted to probability vectors close to their corresponding one-hot labels through a linear transformation (the last FC layer) followed by a softmax function, implicitly with an encoding process of $\boldsymbol v \rightarrow \\W \boldsymbol v \rightarrow \emph{SoftMax}(W \boldsymbol v)$ regularized under the same classification loss, for any $\boldsymbol v \in \mathcal V$ there exists $\delta \rightarrow 0$ satisfying 

 \begin{equation}
 \begin{aligned}
E(\boldsymbol v) &= \emph{S}(W \boldsymbol v) \in \mathcal Q:=\{\boldsymbol p:1 - \boldsymbol p_{y}  < \delta, \\
&\sum_{i=1}^{c}\boldsymbol p_i = 1, \boldsymbol p_i \ge 0\; for\; i = 1,2,...,c \}, 
\label{qdomain}
 \end{aligned}
 \end{equation}

\noindent where $y \in Y := \{1,...,c\}$ represents the ground truth assigning membership to one of $c$ ID classes and $\emph{S}$ denotes the softmax function. $\mathcal Q$ is defined in a normalized space of probability, if the $\delta$ is sufficiently small, each point within $\mathcal Q$ can potentially correspond to the predicted posterior probability of an ID data. Approximately, for $\forall \boldsymbol p \in \mathcal Q, \exists \boldsymbol v \in \mathcal V$ satisfying $E(\boldsymbol v) = \emph{S}(W \boldsymbol v) = \boldsymbol p$, \ie, for $\forall \boldsymbol p \in \mathcal Q, \boldsymbol p \in \mathcal S_{ID}$. Also, since $\mathcal S_{ID} \subset \mathcal Q$ from Eq.\ref{qdomain}, we emphasize that $\mathcal Q$ is such a tractable compact space closely covering $\mathcal S_{ID}$ conditioned on an adequately small $\delta$. 

Fundamentally, to ensure the $\delta$ as small as possible, we suggest not to involve additional loss term like weight decay into the training of $M(\cdot)$ to distract it from generating ID AVs able to utterly minimize the classification loss, \ie, the $\delta$. Thus, the proposed encoder includes only one FC layer and the following softmax, and the regularization term is simply the same classification loss applied for the training of $M(\cdot)$ (here, we assumed cross entropy). 

Since $\mathcal S_{ID} \approx \mathcal Q$ for a small $\delta$, modeling $P(\emph{S}(W \boldsymbol v) \in \mathcal S_{ID})$ is feasible by estimating the probability that $\emph{S}(W \boldsymbol v)_{\overline y}$ is larger than that of an ID data, where $\overline y$ indexes the maximum value. Hence, similar to Eq.\ref{GCCDF}, we express $P(\emph{S}(W \boldsymbol v) \in \mathcal S_{ID})$ as the CDF $\Phi(\emph{S}(W \boldsymbol v)_{\overline y})$ of a ($\mu_0,\sigma_0$) parameterized Gaussian. After employing the proposed two-layer encoder and substituting Eq.\ref{normalized} and \ref{GCCDF} into Eq.\ref{input}, we get the ultimate normality measure 

\begin{small}
\begin{equation}
\begin{aligned}
P(\boldsymbol v \in V) &= \Phi(\emph{S}(\frac{W \boldsymbol v}{T})_{\overline y} | \mu_0,\sigma_0+\epsilon_0) \\
&\cdot \Psi(\Vert \frac{\boldsymbol v}{\Vert \boldsymbol v \Vert} - \frac{D_1(W \boldsymbol v)}{\Vert \boldsymbol v \Vert} \Vert | \mu_1,\sigma_1 + \epsilon_1) \\
&\cdot \Psi(\Vert \frac{W \boldsymbol v}{\Vert W \boldsymbol v \Vert} - \frac{D_2(\emph{S}(\frac{W \boldsymbol v}{T}))}{\Vert \frac{W \boldsymbol v}{T} \Vert} \Vert | \mu_2,\sigma_2+\epsilon_2). \label{score}
\end{aligned}
\end{equation}
\end{small} 

\begin{figure}[h]
\centering
\includegraphics[height=7cm,width=7cm]{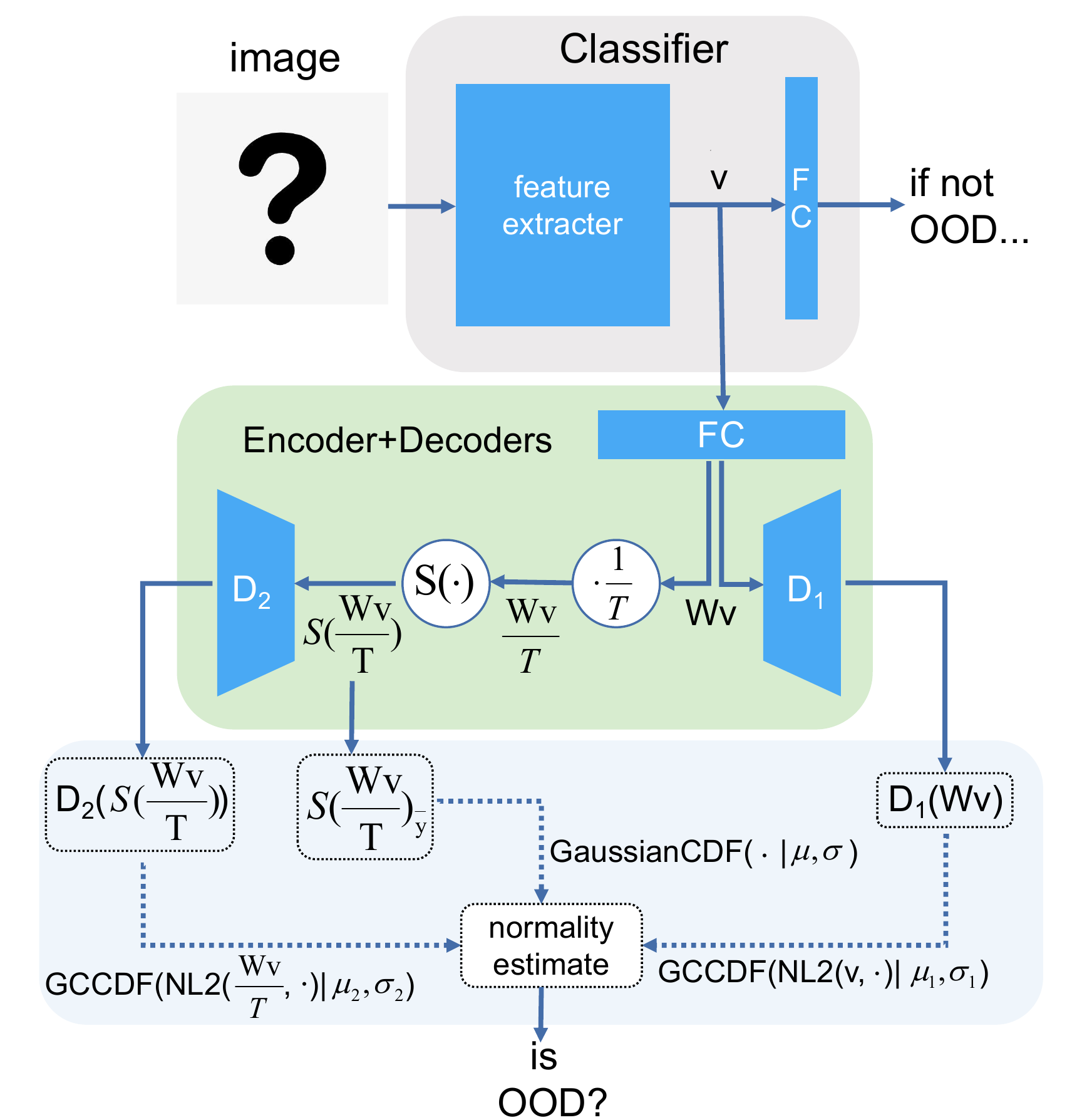}
\caption{The overall framework our method.}
\label{3rd}
\end{figure}

Considering that softmax score in a calibrated probability space can better differentiate OoD from ID \cite{liang2017enhancing}, we apply the temperature scaled logits $\frac{W \boldsymbol v}{T}$ and probability $\emph{S}(\frac{W \boldsymbol v}{T})$ in Eq.\ref{score}. As $W \boldsymbol v \rightarrow \frac{W \boldsymbol v}{T}$ is a bijective mapping, the term to recover information lost in it is ignored. By obtaining $W,D_1,D_2$ and parameters of distributions from the training pipeline summarized in Alg.\ref{training_p}, we could compute the normality score by Eq.\ref{score} for each test image and determine the data abnormality relying on a threshold value that can be set without OoD data (see sec.\ref{threshold}). If an input is recognized as ID, prediction output from the classifier is suggested for its label assignment among known classes. The whole framework is graphed in Fig.\ref{3rd}.

\section{Experiments}

\subsection{Experiment Settings}
{\bf Network and Training Details:} We adopt Dense-BC \cite{huang2017densely} and Wide-ResNet-28-10 \cite{zagoruyko2016wide} as the classifier $M(\cdot)$\\ for AV feature extraction. We do not apply weight decay for both models and use dropout with drop probability of 0.2 for Dense-BC. Other training details are identical to those given in \cite{liang2017enhancing}. Once $M(\cdot)$ has been trained on ID training set, we use the modules before the last FC layer as the feature extractor and further train an encoder along with two decoders for the proposed layerwise semantic reconstruction described in Alg.\ref{training_p}. The FC layer of the proposed encoder is initialized with the parameters of the last FC layer in $M(\cdot)$. Moreover, each decoder is a three-layer FC network as a simple form among the possible variants, for which detailed architecture is listed in the supplementary material. For the training details, Adam solver is applied without Nesterov momentum.The learning rate starts at 1e-4 and decays by a factor of 10 at 50\% and 75\% of total updates. Batch size and epoch number are 128 and 300, respectively. Horizontal flip \begin{strip}
    \begin{minipage}{.764\textwidth}
        \captionof{table}{OoD detection results in CIFAR-10 and CIFAR-100. Our method is compared with three SOTA methods of DAC, ELOC and GODIN. For fair comparison, we use the results of DAC, ELOC and GODIN reported in original papers. If no result is reported as a certain setting, it is marked as $-$. Also, DAC and GODIN did not report experimental results in detection error and AUPR-in. For each evaluation metric, $\uparrow$ means that larger value is better and $\downarrow$ indicates that lower value is better. All values are percentages.}
        \resizebox{\textwidth}{!}{
            \renewcommand{\arraystretch}{0.9}
            \begin{tabular}{ccc|C{0.7cm}C{0.5cm}C{0.8cm}C{0.7cm}|cc|C{0.7cm}C{0.5cm}C{0.8cm}C{0.7cm}|cc|}
                \hline
	\multicolumn{3}{c|}{OoD Dataset}&\multicolumn{4}{c|}{FPR@95\%TPR $\downarrow$}&\multicolumn{2}{c|}{Detection Error $\downarrow$}&\multicolumn{4}{c|}{AUROC $\uparrow$}&\multicolumn{2}{c|}{AUPR-In $\uparrow$}\\
	\hline
	\multicolumn{3}{c|}{}&ELOC&DAC&GODIN&ours&ELOC&ours&ELOC&DAC&GODIN&ours&ELOC&ours\\
	\cline{4-15}
	\multicolumn{1}{c}{\multirow{5}{*}{\rotatebox{90}{\textbf{WRN-28-10}}}} & \multicolumn{1}{c}{\multirow{5}{*}{\rotatebox{90}{CIFAR-10}}} &\multicolumn{1}{c|}{TINc} &\multicolumn{1}{c}{0.8}&\multicolumn{1}{c}{-}&\multicolumn{1}{c}{-}&\multicolumn{1}{c|}{\textbf{0.5}}&\multicolumn{1}{c}{2.2}&\multicolumn{1}{c|}{\textbf{1.9}}&\multicolumn{1}{c}{\textbf{99.8}}&\multicolumn{1}{c}{-}&\multicolumn{1}{c}{-}&\multicolumn{1}{c|}{\textbf{99.8}}&\multicolumn{1}{c}{\textbf{99.8}}&\multicolumn{1}{c|}{\textbf{99.8}} \\
	\multicolumn{1}{c}{}&\multicolumn{1}{c}{}&\multicolumn{1}{c|}{TINr} &\multicolumn{1}{c}{2.9}&\multicolumn{1}{c}{1.9}&\multicolumn{1}{c}{-}&\multicolumn{1}{c|}{\textbf{1.5}}&\multicolumn{1}{c}{3.8}&\multicolumn{1}{c|}{\textbf{3.1}}&\multicolumn{1}{c}{99.4}&\multicolumn{1}{c}{\textbf{99.5}}&\multicolumn{1}{c}{-}&\multicolumn{1}{c|}{\textbf{99.5}}&\multicolumn{1}{c}{99.4}&\multicolumn{1}{c|}{\textbf{99.6}} \\
	\multicolumn{1}{c}{}&\multicolumn{1}{c}{}&\multicolumn{1}{c|}{LSUNc} &\multicolumn{1}{c}{1.9}&\multicolumn{1}{c}{-}&\multicolumn{1}{c}{-}&\multicolumn{1}{c|}{\textbf{0.8}}&\multicolumn{1}{c}{3.2}&\multicolumn{1}{c|}{\textbf{2.0}}&\multicolumn{1}{c}{99.6}&\multicolumn{1}{c}{-}&\multicolumn{1}{c}{-}&\multicolumn{1}{c|}{\textbf{99.8}}&\multicolumn{1}{c}{99.6}&\multicolumn{1}{c|}{\textbf{99.8}} \\
	\multicolumn{1}{c}{}&\multicolumn{1}{c}{}&\multicolumn{1}{c|}{LSUNr} &\multicolumn{1}{c}{0.9}&\multicolumn{1}{c}{1.5}&\multicolumn{1}{c}{-}&\multicolumn{1}{c|}{\textbf{0.5}}&\multicolumn{1}{c}{2.5}&\multicolumn{1}{c|}{\textbf{2.2}}&\multicolumn{1}{c}{\textbf{99.7}}&\multicolumn{1}{c}{99.6}&\multicolumn{1}{c}{-}&\multicolumn{1}{c|}{\textbf{99.7}}&\multicolumn{1}{c}{\textbf{99.7}}&\multicolumn{1}{c|}{\textbf{99.7}} \\
	\multicolumn{1}{c}{}&\multicolumn{1}{c}{}&\multicolumn{1}{c|}{iSUN} &\multicolumn{1}{c}{-}&\multicolumn{1}{c}{-}&\multicolumn{1}{c}{-}&\multicolumn{1}{c|}{2.9}&\multicolumn{1}{c}{-}&\multicolumn{1}{c|}{4.0}&\multicolumn{1}{c}{-}&\multicolumn{1}{c}{-}&\multicolumn{1}{c}{-}&\multicolumn{1}{c|}{99.2}&\multicolumn{1}{c}{-}&\multicolumn{1}{c|}{99.4} \\
	\hline
	\multicolumn{1}{c}{\multirow{5}{*}{\rotatebox{90}{\textbf{WRN-28-10}}}} & \multicolumn{1}{c}{\multirow{5}{*}{\rotatebox{90}{CIFAR-100}}} &\multicolumn{1}{c|}{TINc} &\multicolumn{1}{c}{9.2}&\multicolumn{1}{c}{-}&\multicolumn{1}{c}{-}&\multicolumn{1}{c|}{\textbf{1.5}}&\multicolumn{1}{c}{6.7}&\multicolumn{1}{c|}{\textbf{3.4}}&\multicolumn{1}{c}{98.2}&\multicolumn{1}{c}{-}&\multicolumn{1}{c}{-}&\multicolumn{1}{c|}{\textbf{99.4}}&\multicolumn{1}{c}{98.4}&\multicolumn{1}{c|}{\textbf{99.5}} \\
	\multicolumn{1}{c}{}&\multicolumn{1}{c}{}&\multicolumn{1}{c|}{TINr} &\multicolumn{1}{c}{24.5}&\multicolumn{1}{c}{18.7}&\multicolumn{1}{c}{-}&\multicolumn{1}{c|}{\textbf{6.6}}&\multicolumn{1}{c}{11.6}&\multicolumn{1}{c|}{\textbf{6.4}}&\multicolumn{1}{c}{95.2}&\multicolumn{1}{c}{94.9}&\multicolumn{1}{c}{-}&\multicolumn{1}{c|}{\textbf{98.4}}&\multicolumn{1}{c}{95.5}&\multicolumn{1}{c|}{\textbf{98.0}} \\
	\multicolumn{1}{c}{}&\multicolumn{1}{c}{}&\multicolumn{1}{c|}{LSUNc} &\multicolumn{1}{c}{14.2}&\multicolumn{1}{c}{-}&\multicolumn{1}{c}{-}&\multicolumn{1}{c|}{\textbf{3.7}}&\multicolumn{1}{c}{8.2}&\multicolumn{1}{c|}{\textbf{4.8}}&\multicolumn{1}{c}{97.4}&\multicolumn{1}{c}{-}&\multicolumn{1}{c}{-}&\multicolumn{1}{c|}{\textbf{99.0}}&\multicolumn{1}{c}{97.6}&\multicolumn{1}{c|}{\textbf{99.1}} \\
	\multicolumn{1}{c}{}&\multicolumn{1}{c}{}&\multicolumn{1}{c|}{LSUNr}&\multicolumn{1}{c}{16.5}&\multicolumn{1}{c}{9.2}&\multicolumn{1}{c}{-}&\multicolumn{1}{c|}{\textbf{5.5}}&\multicolumn{1}{c}{9.1}&\multicolumn{1}{c|}{\textbf{5.8}}&\multicolumn{1}{c}{96.8}&\multicolumn{1}{c}{97.9}&\multicolumn{1}{c}{-}&\multicolumn{1}{c|}{\textbf{98.5}}&\multicolumn{1}{c}{97.0}&\multicolumn{1}{c|}{\textbf{98.6}} \\
	\multicolumn{1}{c}{}&\multicolumn{1}{c}{}&\multicolumn{1}{c|}{iSUN}&\multicolumn{1}{c}{-}&\multicolumn{1}{c}{-}&\multicolumn{1}{c}{-}&\multicolumn{1}{c|}{9.0}&\multicolumn{1}{c}{-}&\multicolumn{1}{c|}{7.4}&\multicolumn{1}{c}{-}&\multicolumn{1}{c}{-}&\multicolumn{1}{c}{-}&\multicolumn{1}{c|}{97.9}&\multicolumn{1}{c}{-}&\multicolumn{1}{c|}{98.2} \\
	\hline
	\multicolumn{1}{c}{\multirow{5}{*}{\rotatebox{90}{\textbf{Dense-BC}}}}&\multicolumn{1}{c}{\multirow{5}{*}{\rotatebox{90}{CIFAR-10}}}&\multicolumn{1}{c|}{TINc}&\multicolumn{1}{c}{\textbf{1.2}}&\multicolumn{1}{c}{-}&\multicolumn{1}{c}{6.6}&\multicolumn{1}{c|}{3.7}&\multicolumn{1}{c}{\textbf{2.6}}&\multicolumn{1}{c|}{4.6}&\multicolumn{1}{c}{\textbf{99.7}}&\multicolumn{1}{c}{-}&\multicolumn{1}{c}{98.7}&\multicolumn{1}{c|}{98.9}&\multicolumn{1}{c}{\textbf{99.7}}&\multicolumn{1}{c|}{99.1} \\
	\multicolumn{1}{c}{}&\multicolumn{1}{c}{}&\multicolumn{1}{c|}{TINr} &\multicolumn{1}{c}{\textbf{2.9}}&\multicolumn{1}{c}{-}&\multicolumn{1}{c}{4.2}&\multicolumn{1}{c|}{10.2}&\multicolumn{1}{c}{\textbf{3.8}}&\multicolumn{1}{c|}{7.1}&\multicolumn{1}{c}{\textbf{99.3}}&\multicolumn{1}{c}{-}&\multicolumn{1}{c}{99.1}&\multicolumn{1}{c|}{97.7}&\multicolumn{1}{c}{\textbf{99.3}}&\multicolumn{1}{c|}{98.1} \\
	\multicolumn{1}{c}{}&\multicolumn{1}{c}{}&\multicolumn{1}{c|}{LSUNc} &\multicolumn{1}{c}{3.4}&\multicolumn{1}{c}{-}&\multicolumn{1}{c}{8.5}&\multicolumn{1}{c|}{\textbf{1.7}}&\multicolumn{1}{c}{4.1}&\multicolumn{1}{c|}{\textbf{3.5}}&\multicolumn{1}{c}{99.3}&\multicolumn{1}{c}{-}&\multicolumn{1}{c}{98.3}&\multicolumn{1}{c|}{\textbf{99.4}}&\multicolumn{1}{c}{99.3}&\multicolumn{1}{c|}{\textbf{99.5}} \\
	\multicolumn{1}{c}{}&\multicolumn{1}{c}{}&\multicolumn{1}{c|}{LSUNr} &\multicolumn{1}{c}{\textbf{0.8}}&\multicolumn{1}{c}{-}&\multicolumn{1}{c}{2.4}&\multicolumn{1}{c|}{14.6}&\multicolumn{1}{c}{\textbf{2.2}}&\multicolumn{1}{c|}{7.8}&\multicolumn{1}{c}{\textbf{99.8}}&\multicolumn{1}{c}{-}&\multicolumn{1}{c}{99.4}&\multicolumn{1}{c|}{97.2}&\multicolumn{1}{c}{\textbf{99.8}}&\multicolumn{1}{c|}{97.7} \\
	\multicolumn{1}{c}{}&\multicolumn{1}{c}{}&\multicolumn{1}{c|}{iSUN} &\multicolumn{1}{c}{-}&\multicolumn{1}{c}{-}&\multicolumn{1}{c}{\textbf{2.5}}&\multicolumn{1}{c|}{17.3}&\multicolumn{1}{c}{-}&\multicolumn{1}{c|}{8.8}&\multicolumn{1}{c}{-}&\multicolumn{1}{c}{-}&\multicolumn{1}{c}{\textbf{99.4}}&\multicolumn{1}{c|}{96.7}&\multicolumn{1}{c}{-}&\multicolumn{1}{c|}{97.5} \\
	\hline
	\multicolumn{1}{c}{\multirow{5}{*}{\rotatebox{90}{\textbf{Dense-BC}}}} & \multicolumn{1}{c}{\multirow{5}{*}{\rotatebox{90}{CIFAR-100}}} &\multicolumn{1}{c|}{TINc} &\multicolumn{1}{c}{\textbf{8.3}}&\multicolumn{1}{c}{-}&\multicolumn{1}{c}{12.2}&\multicolumn{1}{c|}{14.8}&\multicolumn{1}{c}{\textbf{6.3}}&\multicolumn{1}{c|}{6.6}&\multicolumn{1}{c}{\textbf{98.4}}&\multicolumn{1}{c}{-}&\multicolumn{1}{c}{97.6}&\multicolumn{1}{c|}{97.1}&\multicolumn{1}{c}{\textbf{98.6}}&\multicolumn{1}{c|}{98.0} \\
	\multicolumn{1}{c}{}&\multicolumn{1}{c}{}&\multicolumn{1}{c|}{TINr} &\multicolumn{1}{c}{20.5}&\multicolumn{1}{c}{-}&\multicolumn{1}{c}{\textbf{6.7}}&\multicolumn{1}{c|}{14.8}&\multicolumn{1}{c}{10.0}&\multicolumn{1}{c|}{\textbf{8.8}}&\multicolumn{1}{c}{96.3}&\multicolumn{1}{c}{-}&\multicolumn{1}{c}{\textbf{98.6}}&\multicolumn{1}{c|}{96.7}&\multicolumn{1}{c}{96.7}&\multicolumn{1}{c|}{\textbf{97.2}} \\
	\multicolumn{1}{c}{}&\multicolumn{1}{c}{}&\multicolumn{1}{c|}{LSUNc} &\multicolumn{1}{c}{14.7}&\multicolumn{1}{c}{-}&\multicolumn{1}{c}{25.0}&\multicolumn{1}{c|}{\textbf{7.8}}&\multicolumn{1}{c}{8.5}&\multicolumn{1}{c|}{\textbf{5.5}}&\multicolumn{1}{c}{97.4}&\multicolumn{1}{c}{-}&\multicolumn{1}{c}{95.3}&\multicolumn{1}{c|}{\textbf{98.0}}&\multicolumn{1}{c}{97.6}&\multicolumn{1}{c|}{\textbf{98.7}} \\
	\multicolumn{1}{c}{}&\multicolumn{1}{c}{}&\multicolumn{1}{c|}{LSUNr} &\multicolumn{1}{c}{16.2}&\multicolumn{1}{c}{-}&\multicolumn{1}{c}{\textbf{6.2}}&\multicolumn{1}{c|}{14.8}&\multicolumn{1}{c}{8.8}&\multicolumn{1}{c|}{\textbf{8.0}}&\multicolumn{1}{c}{97.0}&\multicolumn{1}{c}{-}&\multicolumn{1}{c}{\textbf{98.7}}&\multicolumn{1}{c|}{96.8}&\multicolumn{1}{c}{97.4}&\multicolumn{1}{c|}{\textbf{97.5}} \\
	\multicolumn{1}{c}{}&\multicolumn{1}{c}{}&\multicolumn{1}{c|}{iSUN} &\multicolumn{1}{c}{-}&\multicolumn{1}{c}{-}&\multicolumn{1}{c}{18.6}&\multicolumn{1}{c|}{\textbf{18.0}}&\multicolumn{1}{c}{-}&\multicolumn{1}{c|}{9.9}&\multicolumn{1}{c}{-}&\multicolumn{1}{c}{-}&\multicolumn{1}{c}{\textbf{98.4}}&\multicolumn{1}{c|}{96.1}&\multicolumn{1}{c}{-}&\multicolumn{1}{c|}{96.9} \\
	\hline  
            \end{tabular}}
            \label{benchmark}
    \end{minipage}\quad
    \begin{minipage}[c][5.5cm]{.213\textwidth}
        \centering
        \resizebox{\textwidth}{!}{
            \renewcommand{\arraystretch}{0.90}
        \begin{tabular}{cc|c|}
            \hline
            \multicolumn{2}{|c|}{OoD Dataset}&\multicolumn{1}{c|}{AUROC}\\ 
            
             \hline
             
                \multicolumn{2}{|c|}{}&w/w.o. weight decay\\

                \multicolumn{1}{|c}{\multirow{5}{*}{\rotatebox{90}{CIFAR-10}}}&\multicolumn{1}{c|}{TINc}&\multicolumn{1}{c|}{98.2\quad /\quad98.9}\\
                \multicolumn{1}{|c}{}&\multicolumn{1}{c|}{TINr} &\multicolumn{1}{c|}{96.8\quad /\quad 97.7}\\
                \multicolumn{1}{|c}{}&\multicolumn{1}{c|}{LSUNc} &\multicolumn{1}{c|}{99.2\quad /\quad99.4}\\
                \multicolumn{1}{|c}{}&\multicolumn{1}{c|}{LSUNr} &\multicolumn{1}{c|}{97.0\quad /\quad97.2}\\
                \multicolumn{1}{|c}{}&\multicolumn{1}{c|}{iSUN} &\multicolumn{1}{c|}{96.7\quad /\quad96.7}\\
               
                \hline
                 
                \multicolumn{1}{|c}{\multirow{5}{*}{\rotatebox{90}{CIFAR-100}}} &\multicolumn{1}{c|}{TINc} &\multicolumn{1}{c|}{96.1\quad /\quad97.1}\\
                \multicolumn{1}{|c}{}&\multicolumn{1}{c|}{TINr} &\multicolumn{1}{c|}{94.5\quad /\quad96.7}\\
                \multicolumn{1}{|c}{}&\multicolumn{1}{c|}{LSUNc} &\multicolumn{1}{c|}{98.1\quad /\quad98.0}\\
                \multicolumn{1}{|c}{}&\multicolumn{1}{c|}{LSUNr} &\multicolumn{1}{c|}{93.8\quad /\quad96.8}\\
                \multicolumn{1}{|c}{}&\multicolumn{1}{c|}{iSUN} &\multicolumn{1}{c|}{93.1\quad /\quad96.1}\\
                \hline
        	    \end{tabular}}
            \captionof{table}{AUROC of OoD detection for Dense-BC trained with/without weight decay.}
            \label{weight}
        \includegraphics[width=1.10\textwidth]{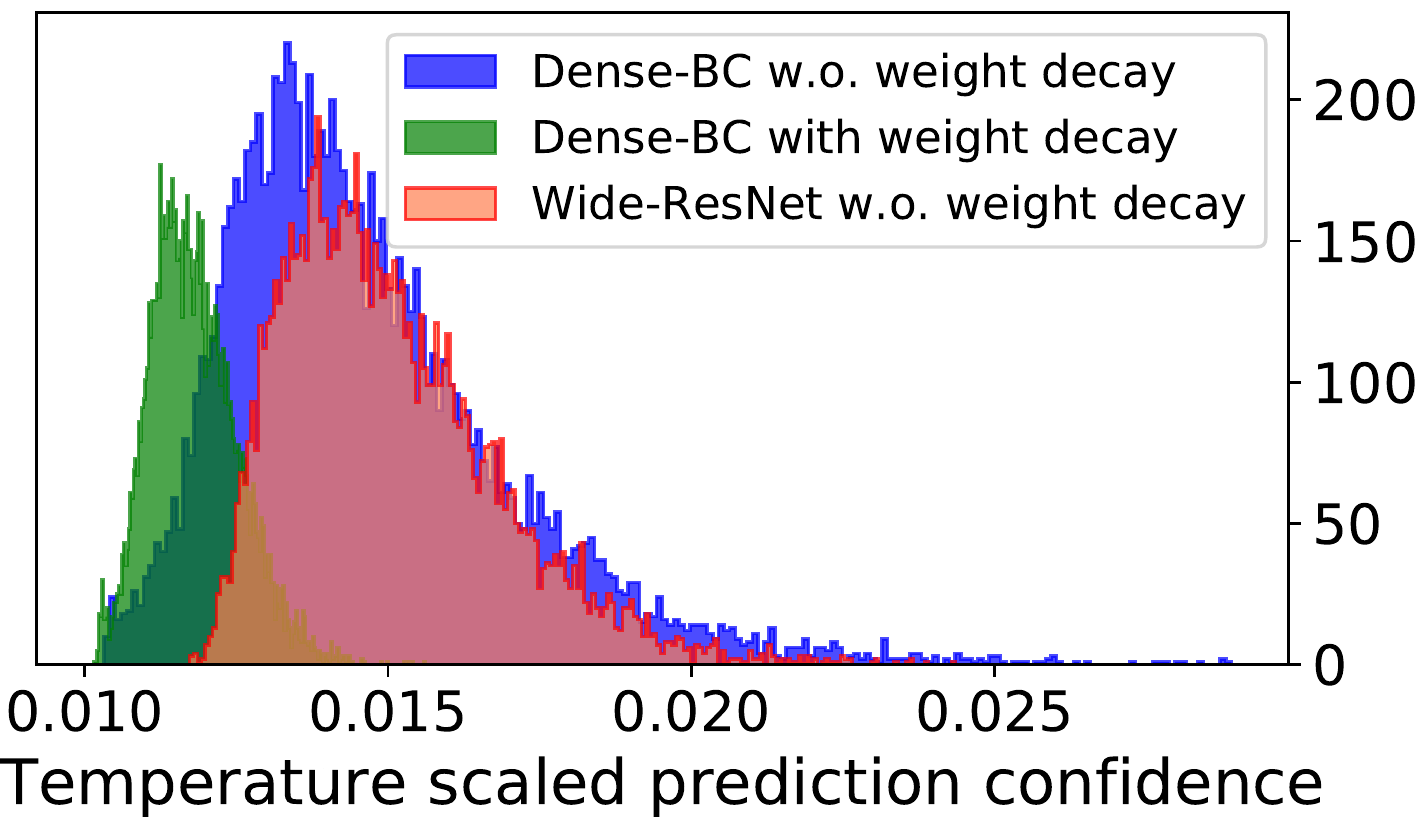}
        \captionof{figure}{The distribution of softmax scores for models in Table.\ref{benchmark} and \ref{weight}. For better distinction we present the temperature-scaled probabilities.}
        \label{sange}
    \end{minipage}
\end{strip}with probability of 0.5 is the only data augmentation, and we do not apply weight decay, dropout, or other refinement tricks. For the hyper parameters, empirically loss weight $\lambda$ is 1, temperature $T$ is set as 100, and $\epsilon_i$, which helps to avoid normality score from collapsing, is applied as 10$\times \sigma_i$.

{\bf Data and Evaluation Metric:} 2000 images with size of 32x32 are randomly held out from the training set of CIFAR-10/100\cite{2009Learning} for validation use especially of the derivation of distribution parameters in Eq.\ref{score}. Other 48000 images are utilized as ID dataset to train the networks. ID test set is the test split of CIFAR-10/100. OoD test sets include TinyImagenet-resize, TinyImagenet-crop, LSUN-resize, LSUN-crop and iSUN stated and released in \cite{liang2017enhancing}. Each test set contains 10000 images except iSUN (8925).

As for the evaluation metrics, since our method does not affect the class assignment of image as long as it is not detected as OoD, we only evaluate model's performance of OoD/ID binary classification based on proposed normality score. Here we simply restate our evaluation metrics: \vspace{-1.1ex} 

\begin{enumerate}
\item {\bf FPR@95\%TPR}: The value of False Positive Rate (FPR) when True Positive Rate (TPR) is 95\%.\vspace{-2.0ex} 
\item {\bf AUROC}: Area Under the Receiver Operating Characteristic curve is the area under the TPR vs. FPR curve. It is a threshold independent metric and a higher value means a better trade off between TPR and FPR. \vspace{-2.0ex} 
\item {\bf AUPR}:Area Under the Precision-Recall curve shows the variation relation between precision and recall, which is also threshold independent. AUPR-in denotes the AUPR where IDs are specified as positive. \vspace{-2.0ex} 
\item {\bf Detection error}: Detection error is computed as the minimal value of 0.5(1-TPR) + 0.5FPR over all possible score thresholds, and could be interpreted as the minimum misclassification probability. 
\end{enumerate}

\begin{table*}[t]
\setlength{\abovecaptionskip}{10pt}
\setlength{\belowcaptionskip}{-0.cm}
\footnotesize
\centering
\caption{The breakdown effect for each effective component of the proposed method on Wide-ResNet-28-10. The ID dataset is chosen as the much challenging CIFAR-100 for better comparison. The leftmost column is the experimental result of original LSA characterized with four latterly alternated components: 1. image: Image is selected as the reconstruction target from which reconstruction error to detect OoD is computed. 2. L2: The distance metric applied to evaluate reconstruction accuracy is L2 distance. 3. AutoReg: An autoregressive regressor is added to compress the latent space. 4.basic: Novelty score is computed by input reconstruction error instead of feature reconstruction errors. It is the framework of vanilla reconstruction autoencoder-based OoD detector introduced in sec.\ref{backrelat}. Each following column to the right represents a model modified from the one next to its left, with main modifications represented on the top, and the component after $-$ is replaced with the component after $+$. The detailed configuration of each is placed in the supplementary material.} 
\label{table1}
\setlength{\tabcolsep}{1.5mm}{
         \renewcommand{\arraystretch}{0.99}
	\begin{tabular}{|cc|cccccc|}
	\hline
	\multicolumn{2}{|c|}{\multirow{2}{*}{Methods}}&\multicolumn{6}{c|}{$\qquad \ \ 1^{st} LSA \qquad \ \ \longrightarrow \qquad 2^{nd} \quad \longrightarrow \quad \ 3^{rd} \ \ \longrightarrow \quad \ \; 4^{th} \quad \ \ \longrightarrow \quad \quad \, 5^{th} \quad \ \ \longrightarrow \quad \, 6^{th}$}\\ 
	\multicolumn{2}{|c|}{}&image,L2,AutoReg,basic&-image+feature&-L2+NL2&-AutoReg+CE&-basic+layerwise&+epsilon\\
	\hline
	\multicolumn{1}{|c|}{\multirow{2}{*}{TINc}}&\multicolumn{1}{c|}{FPR@95\%TPR $\downarrow$}&\multicolumn{1}{c}{42.0}&\multicolumn{1}{c}{99.5}&\multicolumn{1}{c}{5.9}&\multicolumn{1}{c}{3.0}&\multicolumn{1}{c}{\textbf{0.2}}&\multicolumn{1}{c|}{1.5}\\
	\cline{2-2}
	\multicolumn{1}{|c|}{}&\multicolumn{1}{c|}{AUROC $\uparrow$}&\multicolumn{1}{c}{89.2}&\multicolumn{1}{c}{36.7}&\multicolumn{1}{c}{98.6}&\multicolumn{1}{c}{99.1}&\multicolumn{1}{c}{91.3}&\multicolumn{1}{c|}{\textbf{99.4}}\\
	
	\hline
	\multicolumn{1}{|c|}{\multirow{2}{*}{TINr}}&\multicolumn{1}{c|}{FPR@95\%TPR $\downarrow$}&\multicolumn{1}{c}{51.2}&\multicolumn{1}{c}{78.8}&\multicolumn{1}{c}{20.3}&\multicolumn{1}{c}{17.7}&\multicolumn{1}{c}{\textbf{2.5}}&\multicolumn{1}{c|}{6.6}\\
	\cline{2-2}
	\multicolumn{1}{|c|}{}&\multicolumn{1}{c|}{AUROC $\uparrow$}&\multicolumn{1}{c}{89.4}&\multicolumn{1}{c}{80.1}&\multicolumn{1}{c}{95.2}&\multicolumn{1}{c}{96.4}&\multicolumn{1}{c}{90.8}&\multicolumn{1}{c|}{\textbf{98.4}}\\
	
	\hline
	\multicolumn{1}{|c|}{\multirow{2}{*}{LSUNc}}&\multicolumn{1}{c|}{FPR@95\%TPR $\downarrow$}&\multicolumn{1}{c}{55.8}&\multicolumn{1}{c}{100.0}&\multicolumn{1}{c}{4.7}&\multicolumn{1}{c}{4.1}&\multicolumn{1}{c}{\textbf{0.8}}&\multicolumn{1}{c|}{3.7}\\
	\cline{2-2}
	\multicolumn{1}{|c|}{}&\multicolumn{1}{c|}{AUROC $\uparrow$}&\multicolumn{1}{c}{70.0}&\multicolumn{1}{c}{7.5}&\multicolumn{1}{c}{98.1}&\multicolumn{1}{c}{98.7}&\multicolumn{1}{c}{91.3}&\multicolumn{1}{c|}{\textbf{99.0}}\\
	
	\hline
	\multicolumn{1}{|c|}{\multirow{2}{*}{LSUNr}}&\multicolumn{1}{c|}{FPR@95\%TPR $\downarrow$}&\multicolumn{1}{c}{28.2}&\multicolumn{1}{c}{65.6}&\multicolumn{1}{c}{20.0}&\multicolumn{1}{c}{17.7}&\multicolumn{1}{c}{\textbf{1.7}}&\multicolumn{1}{c|}{5.5}\\
	\cline{2-2}
	\multicolumn{1}{|c|}{}&\multicolumn{1}{c|}{AUROC $\uparrow$}&\multicolumn{1}{c}{93.3}&\multicolumn{1}{c}{80.5}&\multicolumn{1}{c}{95.7}&\multicolumn{1}{c}{96.0}&\multicolumn{1}{c}{91.1}&\multicolumn{1}{c|}{\textbf{98.5}}\\
	
	\hline
	\multicolumn{1}{|c|}{\multirow{2}{*}{iSUN}}&\multicolumn{1}{c|}{FPR@95\%TPR $\downarrow$}&\multicolumn{1}{c}{52.5}&\multicolumn{1}{c}{66.4}&\multicolumn{1}{c}{26.0}&\multicolumn{1}{c}{22.4}&\multicolumn{1}{c}{\textbf{2.8}}&\multicolumn{1}{c|}{9.0}\\
	\cline{2-2}
	\multicolumn{1}{|c|}{}&\multicolumn{1}{c|}{AUROC $\uparrow$}&\multicolumn{1}{c}{89.4}&\multicolumn{1}{c}{82.3}&\multicolumn{1}{c}{94.8}&\multicolumn{1}{c}{95.8}&\multicolumn{1}{c}{90.8}&\multicolumn{1}{c|}{\textbf{97.9}}\\
	\hline
	
\end{tabular}}
\label{ablation}
\end{table*}

\subsection{Benchmark Results}

Following the benchmarks given in \cite{liang2017enhancing}, we compare our method against three recent notable SOTA approaches: Ensemble of self supervised Leave-Out Classifiers (ELOC)\cite{vyas2018out}, Generalized ODIN (GODIN)\cite{hsu2020generalized}, and Deep Abstaining Classifier (DAC)\cite{Thulasidasan2021AnEB}. It should be noted that both DAC and ELOC involve data treated as OoD during training. ELOC is an ensemble of multiple classifiers (in Table.\ref{benchmark}, five classifiers), and both ELOC and GODIN apply an input processing strategy requiring an additional inference and time-consuming gradient computation for each image. 

The overall results are presented in Table.\ref{benchmark}. It highlights that for Wide-ResNet-28-10, across all settings, especially for CIFAR-100 and TinyImagenet-resized/LSUN-resized dataset pairs which are commonly considered as challenging cases, our prescription outperforms others with considerable margins. However, for Dense-BC our method delivers less comparable scores in a fraction of the cases.

The performance divergence between Wide-ResNet and Dense-BC is not surprising since the effectiveness of our approach leans on a sufficiently compact latent space to estimate $P(E(\boldsymbol v) \in \mathcal S_{ID})$. Specifically, our method benefits from a smaller $\delta$ in Eq.\ref{qdomain}, \ie, larger prediction confidence of ID samples. For this purpose, as discussed previously, we remove the weight decay of 0.0001 when training the classifiers serving as AV feature extractor to gain performance boost (see Table.\ref{weight}). However, after we remove the weight decay for Dense-BC, the classification accuracy drops drastically from 77\% to 72\% for CIFAR-100 and 95\% to 91\% for CIFAR-10 (those for Wide-ResNet are 78\% to 77\% and 95\% to 94\%, respectively), making it harder for Dense-BC to generate ID AV features able to produce comparably high prediction confidence \cite{hendrycks2016baseline}, as shown in Fig.\ref{sange}. Therefore, our method is much effective when employing classifier less dependent on weight decay as AV feature extractor.

\begin{SCfigure}
\centering
\includegraphics[height=3.5cm,width=5.4cm]{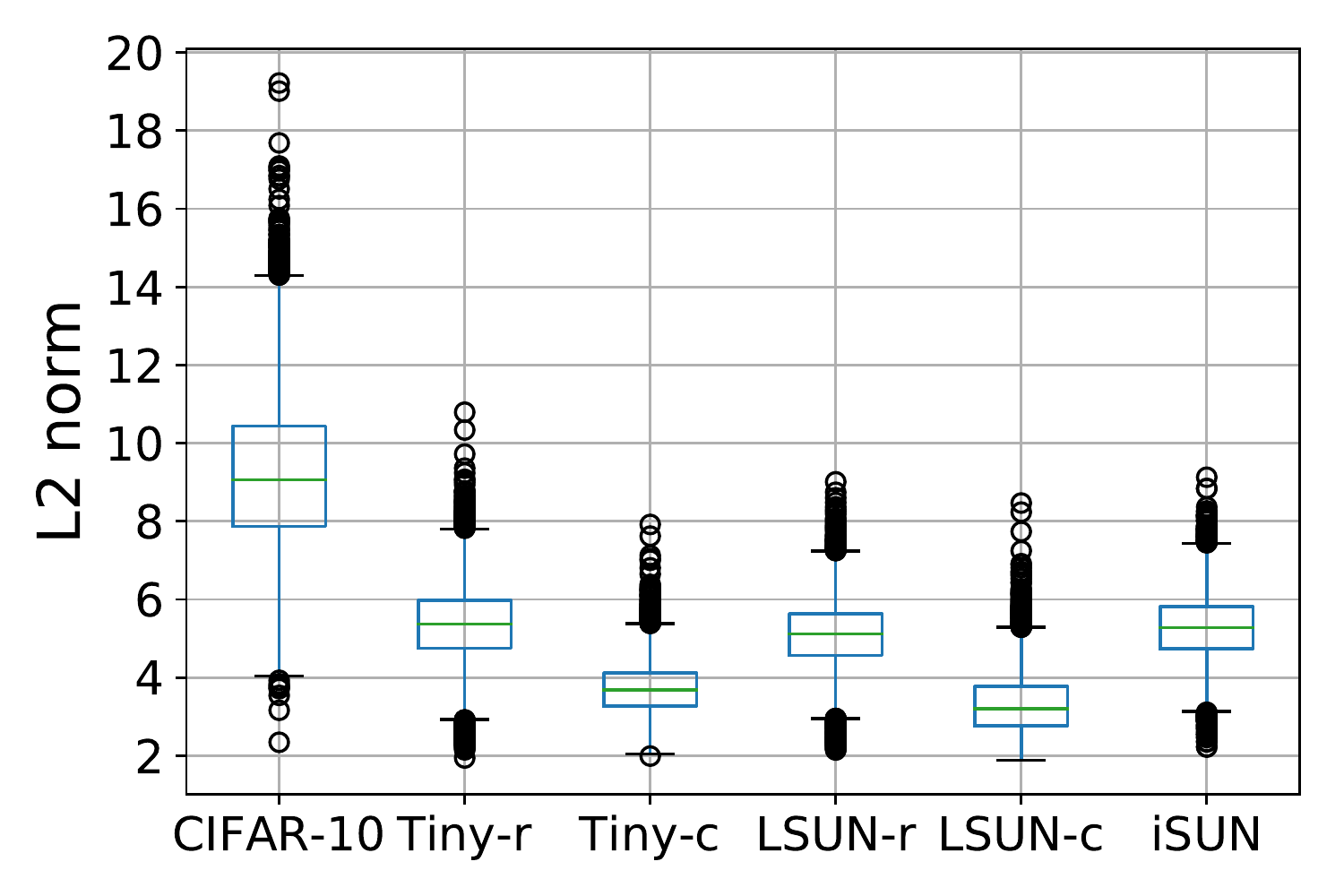}
\caption{The distribution of AV features' L2-norm for CIFAR-10 (ID) test set and various OoD datasets. Features are extracted in the Wide-ResNet.}
\label{thenorm}
\end{SCfigure}

\begin{figure*}[t]
\centering
\subfloat[]{
\begin{minipage}[t]{0.33\linewidth} 
\centering
\includegraphics[width=5.5cm]{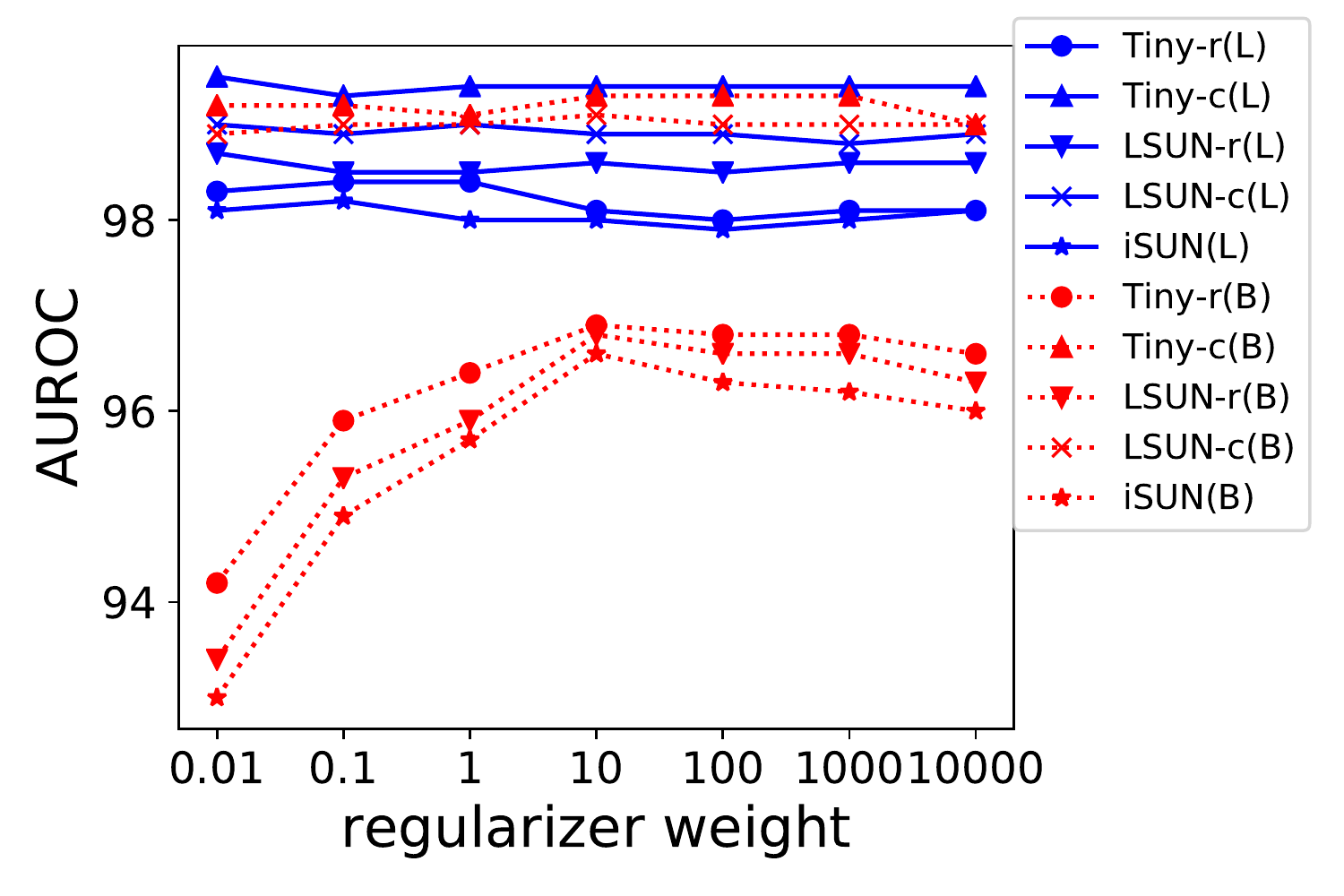}
\end{minipage} \label{layerwise}
}%
\subfloat[]{
\begin{minipage}[t]{0.34\linewidth}
\centering
\includegraphics[width=5.85cm]{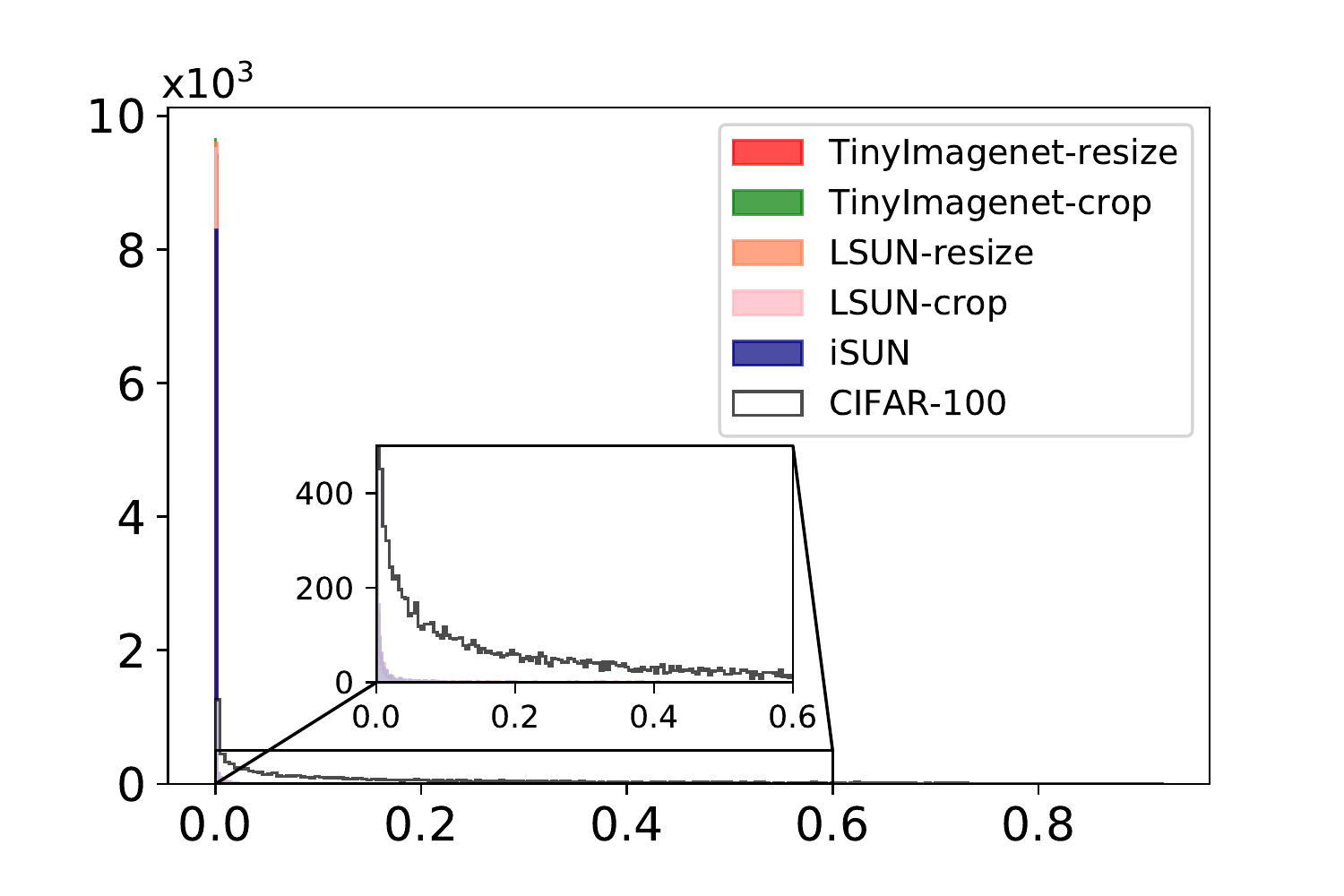}
\end{minipage} \label{without}
}%
\subfloat[]{
\begin{minipage}[t]{0.33\linewidth}
\includegraphics[width=5.30cm]{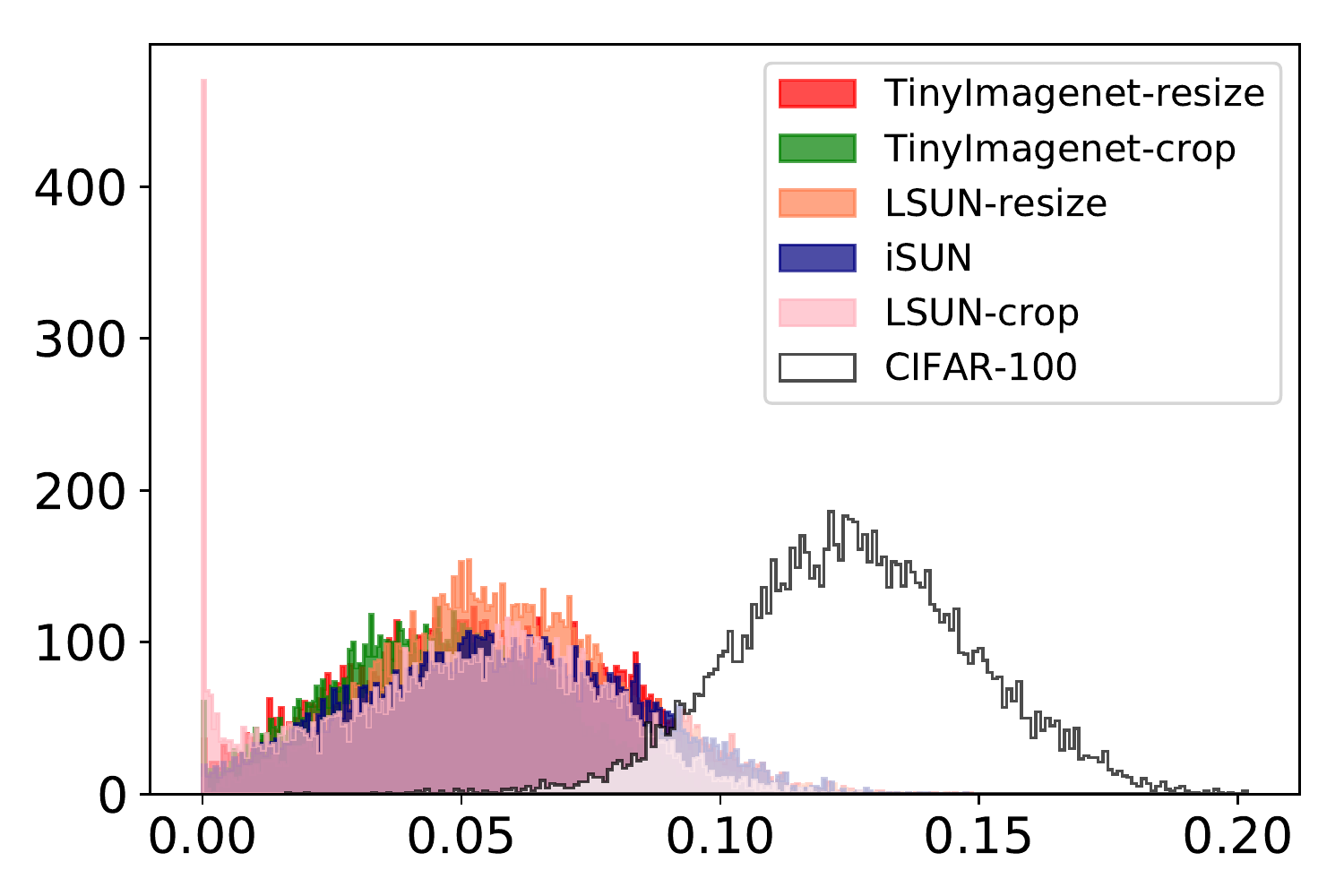}
\end{minipage} \label{with}
}%
\caption{(a): AUROC as a function of the weight of regularizer $\lambda$ for our framework of layerwise reconstruction (blue) vs. the basic framework (red). (b): The distributions of the normality scores computed from Eq.\ref{score} with $\epsilon_i$ applied as 0 and (c) with $\epsilon_i$ applied as $10 \times \sigma_i$.}
\label{scatter}
\end{figure*}

\subsection{Ablation Studies}
To study the effectiveness and characteristics of each technical contribution, we employ the aforementioned latent space autoregression (LSA) \cite{2018Latent}, which is a leading improvement of autoencoder-based OoD detector as a basic method to be gradually modified towards our proposal, and compare the differences produced by each modification. To better discern between OoD detectors, we use CIFAR-100 as the ID dataset in the following ablation studies. The overall comparison is presented in Table.\ref{ablation}. 

\textbf{AV feature as reconstruction target.} We study the effect of changing the reconstruction target from the image to its AV feature. From column $1^{st}$ to $2^{nd}$ in Table.\ref{ablation}, by simply use of the L2 reconstruction error computed from AV feature as decision function, it becomes much difficult to differentiate OoD from ID, even catastrophically. This is expectable since a classifier is inclined to generate AV features with smaller norm on OoD images (Fig.\ref{thenorm}), and such a phenomenon is negative for producing a larger reconstruction error for OoD samples. From column $2^{nd}$ to $3^{rd}$, after altering to using the proposed normalized L2 distance as the evaluation metric of feature reconstruction, the performance in all settings is fostered utterly. That not only proves the effectiveness of the proposed NL2 distance metric, but also validates our point of view that semantic feature could enable autoencoder-based OoD detection better than image. 

\textbf{Cross entropy as latent space regularizer.} Next, we substitute the original encoder and regularizer of LSA with the proposed ones, and apply a normality score based on Eq.\ref{func} (one decoder). The results in column $4^{th}$ indicate that the detector with our encoder scores better in all the cases. Also, compared to the original autoregressor consisting of five masked FC layers with considerable parameters, our regularizer of simple cross entropy, utilizing the nature of ID AV features, is much effective in term of computation.

\textbf{Layerwise reconstruction for certainty decomposition.} With adjusting the basic framework to the one of layerwise reconstruction to detect OoD with the proposed certainty measure, column $5^{th}$ and $6^{th}$ both represent the complete versions of our method. The only difference is that in column $5^{th}$ $\epsilon$ terms in Eq.\ref{score} are not applied. Comparing column $4^{th}$ and $6^{th}$, our framework rooting from the proposed data certainty decomposition performs consistently better with restricted latent space. To further confirm this, we train two sequences of models with the same configuration as that of column $4^{th}$ and $6^{th}$, respectively. They only differ in the loss weight $\lambda$ of regularizer and their performance is compared in Fig.\ref{layerwise}. It can be observed that for models built on the basic framework, by decreasing $\lambda$, the AUROC is improved at first and then degrades rapidly. This reveals that their performance is quite sensitive to the degree of restrictive power imposed over latent space. In contrast, our framework performs both superiority and robustness for a wide range of $\lambda$. Since the optimal weight is hard to set subtly without OoD validation data, our framework excels in both the detection performance and stability of OoD.

\textbf{$\epsilon$ term.} Additionally, column $5^{th}$ and $6^{th}$ indicate that $\epsilon$ terms help to enhance AUROC while increasing the FPR@95\%TPR both to a significant extent. As shown in Fig.\ref{without}, without $\epsilon$ terms, the normality scores of OoD samples mostly condense to 0 while those of ID samples span over a wide space. It is demonstrated that under our framework the OoD sample yielding lower residuals can not have a high prediction confidence and vice versa. Especially for TinyImagenet-crop and LSUN-crop, the model in column $5^{th}$ scores almost touch-the-ceiling in FPR@95\%TPR. Hence, a high-recall threshold set over the output normality score is sufficient to ensure a considerably accurate OoD detection. Compared to the column $4^{th}$, column $6^{th}$ has better results both for AUROC and FPR@95\%TPR. Thus, if a much balanced performance between evaluation metrics is preferred, $\epsilon$ terms can be added to keep ID samples from producing low certainty (Fig.\ref{with}). \label{threshold}

\subsection{Robustness Exploration} \label{avdis}
\begin{figure}[t]
\centering
\subfloat[]{
\begin{minipage}[t]{0.47\linewidth} 
\centering
\includegraphics[width=4.3cm]{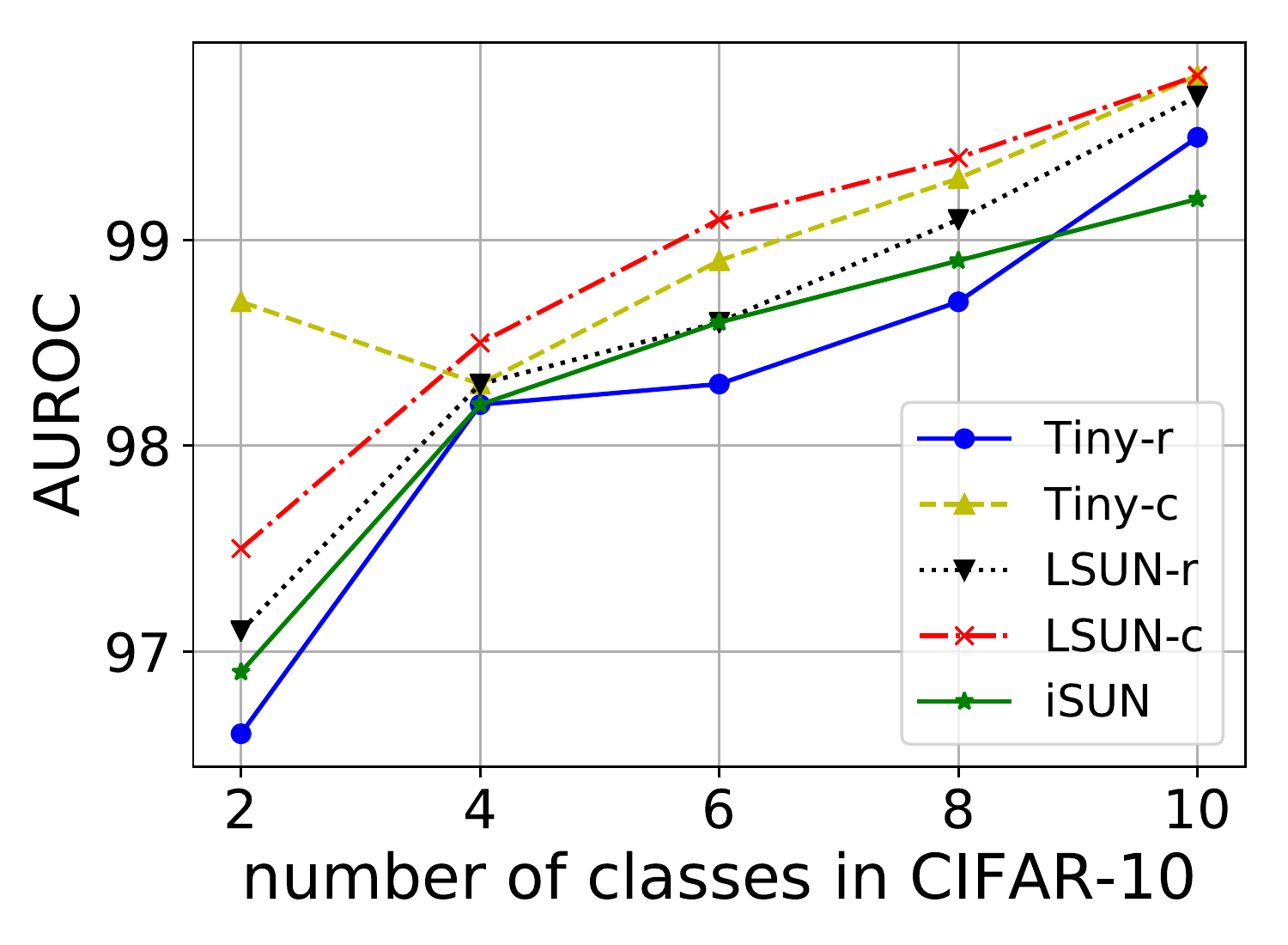}
\end{minipage} \label{num_class}
}%
\subfloat[]{
\begin{minipage}[t]{0.48\linewidth}
\centering
\includegraphics[width=4.3cm]{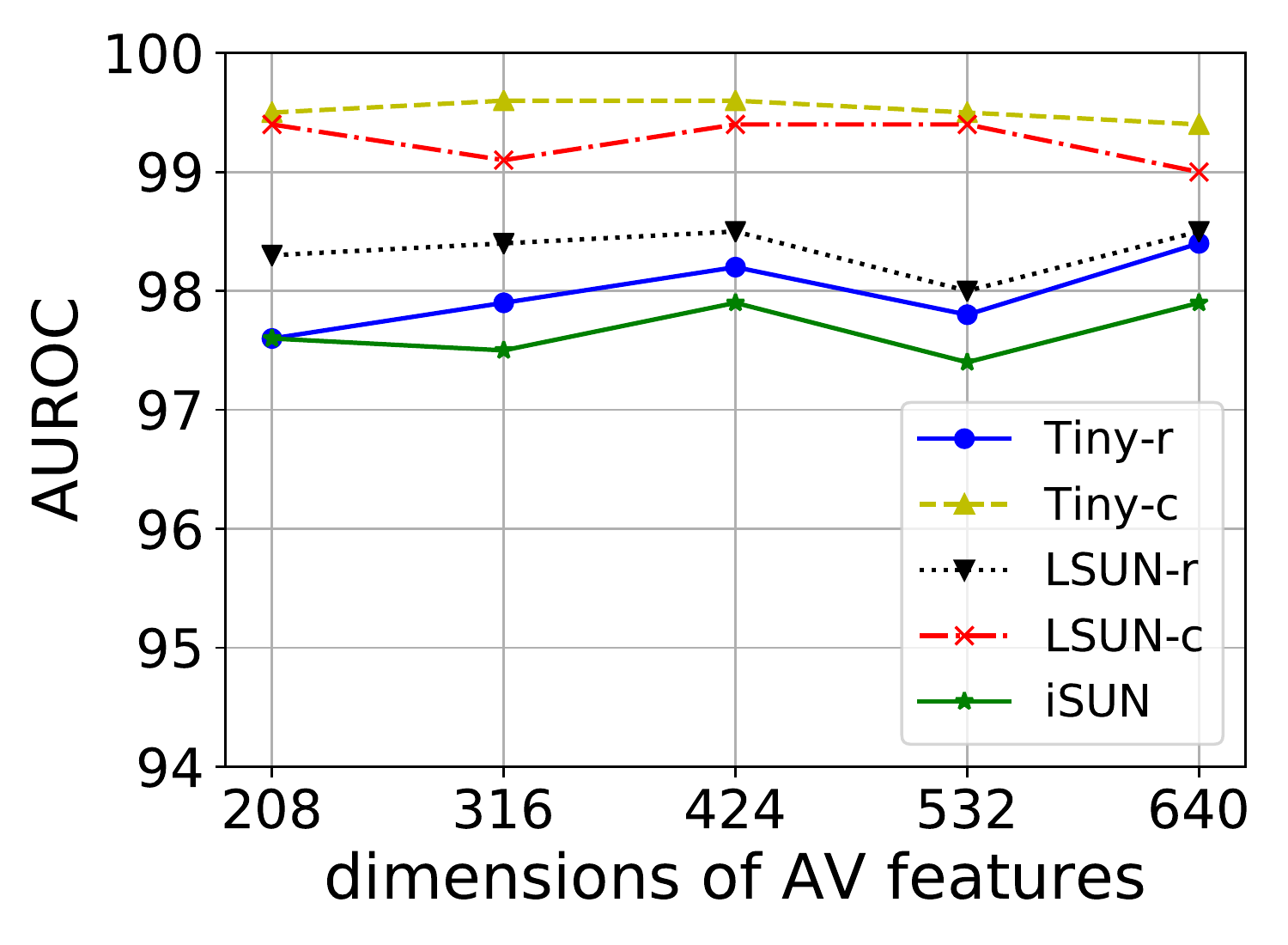}
\end{minipage} \label{av_d}
}%
\caption{AUROC for OoD detection as function of (a): number of ID classes in CIFAR-10 and (b): dimensions of AV features serving as input information of autoencoder. }
\label{scatter}
\end{figure}

\textbf{Number of ID classes.} 
 Since the latent space employed for reconstruction has the same dimensions as the output of classifier, it may be argued that our approach can fail in cases that the number of ID classes is small. For this, we randomly select 2, 4, 6, 8 classes from CIFAR-10 and use them respectively as the ID dataset to assess the robustness of our method to the number of ID classes. As seen in Fig.\ref{num_class}, our method can yield convincible performance on these datasets until the one of 2 classes. One explanation for this is that a latent space with smaller size is conducive to estimating $P(E(\boldsymbol v) \in \mathcal S_{ID})$. Also, the variation of ID AV features decreases while decreasing the number of classes, and thus less expressive power of latent space is required when the number of classes is small.

\textbf{Dimensions of AV feature.} Fig.\ref{av_d} reports a study questioning the robustness of our method to the dimensionality of AV features. For this, we train multiple Wide-ResNet-28-10 differing in the number of channels of the penultimate layer on CIFAR-100 and use them respectively to extract the AV features serving as the input of the autoencoder. The results prove the robustness of our proposal. 

\section{Conclusion}
We propose a novel and effective framework of layerwise semantic reconstruction to enhance reconstruction autoencoder-based OoD detector by retaining the reconstructive power of autoecoder while maximumly compressing its latent space. The comprehensive experiments demonstrate that our method achieves SOTA performance on multi-class OoD detection with merits including orthogonal to classifier, efficient in computation, and OoD data free. Importantly, our theorized perspectives of quadruplet domain translation and data uncertainty decomposition differ from the paradigms of existing methods, heuristically exploring rooms for future work.

\section{Acknowledgements}
We thank Xiaohong Zhou from Chinese Academy of Sciences (CAS), Xiang Zhang from Purdue University and Yiqun Liu from OPPO for partly supporting this work.

{\small
\bibliographystyle{ieee_fullname}
\bibliography{refer}
}

\end{document}